\documentclass[nohyperref]{article}

\usepackage[dvipsnames,table]{xcolor}
\usepackage{nicefrac}
\usepackage{microtype}
\usepackage{graphicx}
\usepackage{tikz}
\usepackage[accepted]{icml2022}
\usepackage{amsmath,amsfonts,bm}
\usepackage{bookmark}
\usepackage{inconsolata}
\usepackage{tcolorbox}
\usepackage{hyperref}
\usepackage{bbm}
\usepackage[nameinlink,capitalize]{cleveref}
\usepackage{enumitem}
\usepackage{multirow}
\usepackage{mathtools}
\usepackage[hyperpageref]{backref}
\usepackage{url}

\definecolor{bg}{rgb}{0.95, 0.95, 0.95}

\newcommand{\code}[1]{\colorbox{bg}{\texttt{#1}}}

\newcommand{\papertitle}{Fast Finite Width Neural Tangent Kernel}

\hypersetup{
    unicode=False,
    pdftoolbar=True,
    pdfmenubar=True,
    pdffitwindow=False,
    pdfstartview={FitH},
    pdftitle={\papertitle},
    pdfauthor={Roman Novak},
    pdfsubject={\papertitle},
    pdfkeywords={Fast, Empirical, Finite Width, Neural Tangent Kernel, NTK, JAX, Jacobian, Algorithm, Software, Automatic Differentiation, AD},
    pdfnewwindow=True,
    colorlinks=True,
    breaklinks=True,
    linkcolor=Bittersweet,
    citecolor=RawSienna,
    filecolor=magenta,
    urlcolor=blue,
}

\title{\papertitle}
\icmltitlerunning{\papertitle}

\newcommand{\email}[1]{\tt\small\href{mailto:#1@google.com}{#1@google.com}}

\newcommand{\sref}[1]{\S\ref{#1}}

\renewcommand{\o}[0]{{\color{blue}\textbf{O}}}  
\newcommand{\oi}[0]{{\color{blue}o}}  

\newcommand{\p}[0]{{\color{red}\textbf{P}}}   
\renewcommand{\t}[0]{{\color{Fuchsia}\textbf{L}}}   
\newcommand{\ti}[0]{{\color{Fuchsia}l}}   

\renewcommand{\c}[0]{{\color{NavyBlue}\textbf{C}}}   
\newcommand{\ci}[0]{{\color{NavyBlue}c}}   

\renewcommand{\d}[0]{{\color{teal}\textbf{D}}} 
\newcommand{\f}[0]{{\color{CadetBlue}\textbf{F}}} 

\renewcommand{\j}[0]{{\color{RoyalPurple}\textbf{J}}} 

\newcommand{\y}[0]{{\color{olive}\textbf{Y}}} 

\renewcommand{\l}[0]{{\color{Brown}\textbf{K}}}   
\newcommand{\li}[0]{{\color{Brown}k}}   

\newcommand{\n}[0]{{\color{purple}\textbf{N}}}   
\renewcommand{\ni}[0]{{\color{purple}n}}   

\newcommand{\w}[0]{{\color{cyan}\textbf{W}}}   

\newcommand{\nt}[0]{\Theta_\theta^f\left[\ti, \li_1, \li_2\right]}

\newcommand{\fp}[0]{\textbf{FP}}

\definecolor{sdc}{HTML}{2ca02c}
\definecolor{jcc}{HTML}{1f77b4}
\definecolor{ntvpc}{HTML}{ff7f0e}
\definecolor{jacc}{HTML}{d62728}

\newcommand{\sd}[0]{\hyperref[sec:str_derivatives]{\color{sdc}Structured derivatives}}
\newcommand{\ntvp}[0]{\hyperref[sec:implicit]{\color{ntvpc}NTK-vector products}}
\newcommand{\jc}[0]{\hyperref[sec:vanilla]{\color{jcc}Jacobian contraction}}
\newcommand{\jac}[0]{\hyperref[sec:jacobian]{\color{jacc}Jacobian}}
\newcommand{\xla}[0]{\href{https://www.tensorflow.org/xla}{XLA}}

\newcommand{\cbd}[0]{\hyperref[sec:constant_block_diagonal]{Constant block-diagonal}}
\newcommand{\bd}[0]{\hyperref[sec:block_diagonal]{Block-diagonal}}
\newcommand{\bt}[0]{\hyperref[sec:block_tiled]{Block-tiled}}
\newcommand{\obt}[0]{\hyperref[sec:output_block_tiled]{Output block-tiled}}
\newcommand{\ibt}[0]{\hyperref[sec:input_block_tiled]{Input block-tiled}}

\begin{document}

\twocolumn[
    \icmltitle{\papertitle}
    
    \icmlsetsymbol{equal}{*}
    
    \begin{icmlauthorlist}
        \icmlauthor{Roman Novak}{g}
        \icmlauthor{Jascha Sohl-Dickstein}{g}
        \icmlauthor{Samuel S. Schoenholz}{g}
    \end{icmlauthorlist}
    
    \icmlaffiliation{g}{Google Brain, \,\,Mountain View, California, United States}
    
    \icmlcorrespondingauthor{Roman Novak}{{\email{romann}}}
    
    \icmlkeywords{Fast, Empirical, Finite Width, Neural Tangent Kernel, NTK, JAX, Jacobian, Algorithm, Software, Automatic Differentiation, AD}
    
    \vskip 0.3in
]

\printAffiliationsAndNotice{}

\begin{tikzpicture}[remember picture, overlay]
    \node[opacity=0.1, shift={(-0.3cm,11.4134cm)}] at (current page.center){
            \includegraphics[scale=0.049]{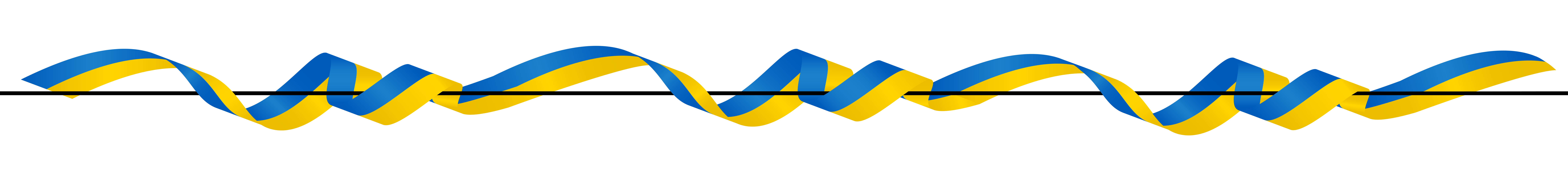}
    };
\end{tikzpicture}

\begin{abstract}
    The Neural Tangent Kernel (NTK), 
    defined as $\Theta_\theta^f(x_1, x_2) = \left[\partial f(\theta, x_1)\big/\partial \theta\right] \left[\partial f(\theta, x_2)\big/\partial \theta\right]^T$ where $\left[\partial f(\theta, \cdot)\big/\partial \theta\right]$ is a neural network (NN) Jacobian, has emerged as a central object of study in deep learning. In the infinite width limit, the NTK can sometimes be computed analytically and is useful for understanding training and generalization of NN architectures. At finite widths, the NTK is also used to better initialize NNs, compare the conditioning across models, perform architecture search, and do meta-learning. Unfortunately, the finite width NTK is notoriously expensive to compute, which severely limits its practical utility. We perform the first in-depth analysis of the compute and memory requirements for NTK computation in finite width networks. 
    Leveraging the structure of neural networks, we further propose two novel algorithms that change the {\em exponent} of the compute and memory requirements of the finite width NTK, dramatically improving efficiency.
    Our algorithms can be applied in a black box fashion to {\em any} differentiable function, including those implementing neural networks. We open-source our implementations within the Neural Tangents package \citep{neuraltangents2020} at \href{https://github.com/google/neural-tangents}{github.com/google/neural-tangents}.
\end{abstract}

\nocite{*}

\section{Introduction}

    The past few years have seen significant progress towards a theoretical foundation for deep learning. Much of this work has focused on understanding the properties of random functions in high dimensions. One significant line of work~\citep{neal,lee2018deep,matthews2018,borovykh2018gaussian,garriga2018deep,novak2018bayesian,yang2019scaling,hron2020,hron2020exact,Hu2020InfinitelyWG} established that in the limit of infinite width, randomly initialized Neural Networks (NNs) are Gaussian Processes (called NNGPs). Building on this development, \citet{Jacot2018ntk} showed that in function space the dynamics under gradient descent could be computed analytically using the so-called Neural Tangent Kernel (NTK) and \citet{lee2019wide} showed that wide neural networks reduce to their linearizations in weight space throughout training. A related set of results~\citep{belkin2019reconciling, spigler2019jamming} showed that the ubiquitous bias-variance decomposition breaks down as high-dimensional models enter the so-called interpolating regime. Together these results describe learning in the infinite width limit and help explain the impressive generalization capabilities of NNs.
    
    Insights from the wide network limit have had significant practical impact. The conditioning of the NTK has been shown to significantly impact trainability and generalization in NNs~\citep{schoenholz2016deep, xiao18a, xiao2019disentangling}. This notion inspired initialization schemes like Fixup~\citep{zhang2019fixup}, MetaInit~\citep{dauphin2019metainit}, and Normalizer Free networks~\citep{brock2021characterizing, brock2021high}, and has enabled efficient neural architecture search~\citep{park2020towards,chen2021vision}. The NTK has additionally given insight into a wide range of phenomena such as: behavior of Generative Adversarial Networks~\citep{franceschi2021neural}, neural scaling laws~\citep{bahri2021explaining}, and neural irradiance fields~\citep{tancik2020fourfeat}. Kernel regression using the NTK has further enabled strong performance on small datasets~\citep{Arora2020Harnessing}, and applications such as approximate inference \citep{khan2019approximate}, dataset distillation~\citep{nguyen2020dataset, nguyen2021dataset}, and uncertainty prediction~\citep{he2020bayesian, adlam2020exploring}.
    
    Despite the significant promise of theory based on the NTK, computing the NTK in practice is challenging. In the infinite width limit, the NTK can sometimes be computed analytically. However, the infinite-width kernel remains intractable for many architectures and finite width corrections are often important to describe actual NNs used in practice (see \sref{sec:finite_vs_inf} for detailed discussion). The NTK matrix can be computed for finite width networks as the outer product of Jacobians using forward or reverse mode automatic differentiation (AD),
    \begin{equation}\label{eq:ntk_outer_product}
        \underbrace{\Theta_\theta^f(x_1, x_2)}_{\o\times \o} \coloneqq 
        \underbrace{\left[\partial f(\theta, x_1)\big/\partial \theta\right]}_{\o \times \p} 
        \underbrace{\left[\partial f(\theta, x_2)\big/\partial \theta\right]^T}_{\p\times \o}
        ,
    \end{equation}
    where $f$ is the forward pass NN function producing outputs in $\mathbb{R}^{\o}$, $\theta\in\mathbb{R}^{\p}$ are all trainable parameters, and $x_1$ and $x_2$ are two  inputs to the network. If inputs are batches of sizes $\n_1$ and $\n_2$, the NTK is an $\n_1\o\times\n_2\o$ matrix.
    
    Unfortunately, evaluating \cref{eq:ntk_outer_product} is often infeasible due to time and memory requirements. For modern machine learning tasks $\o$ is often greater (sometimes much greater) than $1000$ (e.g. for ImageNet \citep{deng2009imagenet}), while even modestly sized models feature tens of millions of parameters, or $\p\sim 10^7$. This makes both storing ($\left[\n_1 + \n_2\right]\o\p$ memory) and contracting ($\left[\n_1\n_2\right]\o^2\p$ time) the Jacobians in \cref{eq:ntk_outer_product} very costly. The theoretical importance of the NTK together with its prohibitive computational costs implies that performance improvements  will unlock impactful novel research. 
    
    We perform the first in-depth analysis of the compute and memory requirements for the NTK as in \cref{eq:ntk_outer_product}. Noting that forward and reverse mode AD are two extremes of a wide range of AD strategies~\citep{naumann2004optimal, naumann2008optimal}, we explore other methods for computing the NTK leveraging the structure of NNs used in practice. We propose two novel methods for computing the NTK that exploit different orderings of the computation. We describe the compute and memory requirements of our techniques in fully-connected (FCN) and convolutional (CNN) settings, and show that one is asymptotically more efficient in both settings. We compute the NTK over a wide range of NN architectures and demonstrate that these improvements are robust in practice. We open-source our implementations as general-purpose JAX\footnote{Our algorithms are framework-agnostic, but implementation in JAX is easier, as described in \sref{sec:jax}. We also provide instructions for implementation in other frameworks like Tensorflow \citep{abadi2016tensorflow} and PyTorch \citep{pytorch} in \sref{sec:jax}.} \citep{jax2018github} function transformations.

\section{Related Work}\label{sec:related}

    The finite width NTK (denoted simply NTK throughout this work\footnote{See \sref{sec:finite_vs_inf} for  a comparison between the finite and infinite width settings.}) has been used extensively in many recent works, but to our knowledge implementation details and compute costs were rarely made public. Below we draw comparison to some of these works, but we stress that it only serves as a sanity check to make sure our contribution is valuable relative to the scale of problems that have been attempted. None of these works had efficient NTK computation as their central goal.
    
    In order to compare performance of models based on the NTK and the infinite width NTK, \citet[Table 2]{arora2019on} compute the NTK of up to 20-layer, 128-channel CNN in a binary CIFAR-2 classification setting. In an equivalent setting with the same hardware (NVIDIA V100), we are able to compute the NTK of a 2048-channel CNN, i.e. a network with at least 256 times more parameters.
    
    To demonstrate the stability of the NTK during training for wide networks, \citet[Figure S6]{lee2019wide} compute the NTK of up to 3-layer $2^{12}$-wide or 1-layer $2^{14}$-wide FCNs. In the same setting with the same hardware (NVIDIA V100), we can reach widths of at least $2^{14}$ and $2^{18}$ respectively, i.e. handle networks with 4 to 16 times more parameters. 
    
    To investigate convergence of a WideResNet WRN-28-$k$ \citep{zagoruyko2016wide} to its infinite width limit, \citet[Figure 2]{neuraltangents2020} evaluate the NTK of this model with widening factor $k$ up to 32. In matching setting and hardware, we are able to reach a widening factor of at least 64, i.e. work with models at least 4 times larger.
    
    To meta-learn NN parameters for transfer learning in a MAML-like \citep{pmlr-v70-finn17a} setting, \citet[Table 7]{zhou2021metalearning} replace the inner training loop with NTK-based inference. They use up to 5-layer, 200-channel CNNs on MiniImageNet \citep{Oreshkin2018TADAMTD} with scalar outputs and batch size 25. In same setting we achieve at least $512$ channels, i.e. support models at least 6 times larger.
    
    \citet[\S 4.1]{park2020towards} use the NTK to predict the generalization performance of architectures in the context of Neural Architecture Search \citep[NAS]{zoph2017neural}; however, the authors comment on its high computational burden and ultimately use a different proxy objective. In another NAS setting, \citet[\S 3.1.1]{chen2020tenas} use the condition number of NTK to predict a model's trainability. \citet[Table 1]{chen2021vision} use the NTK to evaluate the trainability of several ImageNet models such as ResNet 50/152 \citep{he2016deep}, Vision Transformer \citep{dosovitskiy2021an} and MLP-Mixer \citep{tolstikhin2021mlpmixer}. However, due to the prohibitive computational cost, in all of these cases the authors only evaluate a pseudo-NTK, i.e. the NTK of a scalar-valued function,\footnote{Precisely, computing the Jacobian only for a single logit or the sum of all 1000 class logits. The result is not the full NTK, but rather a single diagonal block or the sum of its 1000 diagonal blocks (the finite width NTK is a dense matrix, not block-diagonal).} which impacts the quality of the respective trainability/generalization proxy. 
    
    By contrast, in this work we can compute the full $1000 \times 1000$ ($1000$ classes) NTK for the same models, i.e. perform a task 1000 times more costly. 
    
    Finally, we remark that in all of the above settings, scaling up by increasing width or by working with the true NTK (vs the pseudo-NTK) should lead to improved downstream task performance due to a better infinite width/linearization approximation or a higher-quality trainability/generalization proxy respectively, which makes our work especially relevant to modern research.

\section{Algorithms for Efficient NTK Computation}\label{sec:main_text}
    
    We now describe our algorithms for fast NTK computation.
    
    In \sref{sec:preliminaries} we cover preliminaries. We begin by introducing notation used throughout the paper (\sref{sec:notation}). We then (\sref{sec:jvps_vjps_main}) describe primitive building blocks of AD including the Jacobian-vector products (JVP) and vector-Jacobian products (VJP) that correspond to forward and reverse mode AD respectively before discussing the Jacobian (\sref{sec:jacobian}). 

    In \sref{sec:vanilla} we apply the above tools to describe the computational complexity of the baseline approach to computing the NTK that is used in most (likely all) prior works.
    
    In \sref{sec:implicit} and \sref{sec:str_derivatives} we present our two algorithms that each enable accelerating the computation by orders of magnitude 
    in different ways.\vspace{-0.06cm}

    \subsection{Preliminaries}\label{sec:preliminaries}
    
        \subsubsection{Notation}\label{sec:notation} 
        
            Consider a NN $f(\theta, x)\in\mathbb{R}^\o$ with $\o$ outputs (e.g. class logits) per input $x$ and a total number $\p$ of trainable parameters $\theta = \textrm{vec}\left[\theta^0,\dots,\theta^\t\right]$, with each $\theta^\ti$ of size $\p^\ti$, $\p=\sum_{\ti=0}^{\t}\p^{\ti}$. Also assume the network has $\l$ intermediate \href{https://jax.readthedocs.io/en/latest/notebooks/How_JAX_primitives_work.html}{primitive} outputs $y^{\li}$ of size $\y^\li$ each (for example, activations or pre-activations), and let $\y=\sum_{\li=1}^{\l}\y^\li$ be the total size of the outputs (see \cref{fig:fcn_legend} and \cref{fig:jvp_vjp}). The NTK is
        	\begin{align}\label{eq:ntk_def_main}
            	\underbrace{\Theta_\theta^f}_{\o\times\o} \coloneqq 
            	&\underbrace{\left[{\partial f\left(\theta, x_1\right)}\big/{\partial\theta}\right]}_{\o \times \p}\underbrace{\left[\partial f\left(\theta, x_2\right)\big/{\partial\theta}\right]^T}_{\p\times\o} =\\ 
            	\sum_{\ti=0}^{\t}&
            	\underbrace{\left[{\partial f\left(\theta, x_1\right)}\big/{\partial\theta^\ti}\right]}_{\o\times\p^\ti}\underbrace{\left[{\partial f\left(\theta, x_2\right)}\big/{\partial\theta^\ti}\right]^T}_{\p^\ti\times\o}
            	.
            \end{align}
        	We denote $\fp$ to be the (time or memory, depending on context) cost of a single forward pass $f(\theta, x)$. For memory, we exclude the cost of storing all $\p$ weights, but rather define it to be the cost of evaluating $f$ one primitive $y^\li$ at a time. This gives a memory cost of at most $\mathcal{O}\left(\max_{\li} \y^{\li} + \max_{\ti}\p^{\ti}\right)$, which we denote as simply $\y^\li + \p^\ti$.\footnote{To declutter notation throughout this work, in time and memory complexity expressions, we (1) omit the $\mathcal{O}$ symbol, and (2) imply taking the maximum over any free index.} Finally, we  will consider $x_1$ and $x_2$ to be batches of $\n$ inputs each, in which case the NTK will be an $\n\o\times\n\o$ matrix, obtained by computing \cref{eq:ntk_def_main} for each pair of inputs. See \sref{sec:glossary} for glossary.\vspace{-0.06cm}

        \subsubsection{Jacobian-vector Products (JVP) and Vector-Jacobian Products (VJP)}\label{sec:jvps_vjps_main}
            
            Following \citet{maclaurin2015autograd} we define
            \begin{align}\label{eq:jvp_vjp_main}
                \textrm{JVP}^f_{\left(\theta, x\right)}&\colon 
                \theta_t \in\mathbb{R}^{\p} \mapsto \left[{\partial f\left(\theta, x\right)}\big/{\partial\theta}\right]\theta_t \in\mathbb{R}^{\o}; \\
                \textrm{VJP}^f_{\left(\theta, x\right)}&\colon 
                f_c \in\mathbb{R}^{\o} \mapsto \left[{{\partial f\left(\theta, x\right)}\big/{\partial\theta}}\right]^T f_c \in\mathbb{R}^{\p}
                .
            \end{align}
            The JVP can be understood as pushing forward a tangent vector $\theta_t$ in weight space to a tangent vector in the space of outputs; by contrast the VJP pulls back a cotangent vector $f_c$ in the space of outputs to a cotangent vector in weight space. These elementary operations enable forward and reverse mode AD respectively and serve as a basis for typical AD computations such as gradients, Jacobians, Hessians, etc.
            
            The time cost of both is comparable to \fp~(see \sref{sec:jvp_vjp_costs} and \citet{evaluating_derivatives}). The memory cost of a JVP is \fp~as well (i.e. $\y^\li + \p^\ti$), while the memory cost of a VJP is generally $\y + \p$, since it requires storing all $\l$ intermediate primitive outputs for efficient backprop and all $\t$ output cotangents. However, for the purpose of computing the NTK, we never need to store the whole Jacobian $\partial f/\partial \theta$, but only individual cotangents like $\partial f/\partial\theta^\ti$ to compute the sum in \cref{eq:ntk_def_main} layer-by-layer. Hence we consider VJP to cost $\y + \p^\ti$ memory. To summarize, for a batch of $\n$ inputs,
            \begin{tcolorbox}
                \begin{itemize}[leftmargin=*]
                    \item JVP costs $\n\left[\fp\right]$ time; $\n\left[\y^\li\right] + \p$ memory.
                    \item VJP costs $\n\left[\fp\right]$ time; $\n\left[\y + \p^\ti\right] + \p$ memory.
                \end{itemize}
            \end{tcolorbox}\vspace{-0.1cm}

        \subsubsection{\texorpdfstring{\jac}{Jacobian}}\label{sec:jacobian}
        
            The reverse mode Jacobian $\partial f/\partial \theta$ is  computed via $\o$ VJP calls on rows of the identity matrix $I_\o \in \mathbb{R}^{\o\times \o}$, i.e. 
            \begin{equation}
                \left[{\partial f\left(\theta, x\right)\big/\partial\theta}\right]^T = 
                \left[{\partial f\left(\theta, x\right)\big/\partial\theta}\right]^T I_\o \in \mathbb{R}^{\p\times \o},
            \end{equation}
            and therefore costs $\o\left[\textrm{VJP}\right]$ time and memory apart from parameters and primitive outputs that can be reused across VJPs. Therefore, for a batch of $\n$ inputs,
            \begin{tcolorbox}
                \jac~costs 
                $\n\o\left[\fp\right]$ time; 
                $\n\o\left[\y^\li + \p^\ti\right] + \n\y + \p$ memory.
            \end{tcolorbox}\vspace{-0.1cm}

        \subsection{\texorpdfstring{\jc}{Jacobian Contraction} -- the Baseline}\label{sec:vanilla}
            
            This baseline method of evaluating the NTK consists in computing the Jacobians $\partial f/\partial \theta$ and contracting them as in \cref{eq:ntk_def_main}. The contraction costs $\n^2\o^2\p$ time and $\n^2\o^2 + \n\o\p^\ti$ memory, to store the result $\Theta_\theta^f$ and individual layer-by-layer cotangents $\partial f / \partial \theta^\ti$. Including the cost of computing the cotangents via the batch \jac~$\partial f/\partial \theta = \left[\partial f / \partial \theta^0,\dots,\partial f / \partial \theta^\t\right]$ from \sref{sec:jacobian} we arrive at 
            \begin{tcolorbox}
                \jc~costs 
                $\n\o\left[\fp\right] + \n^2\o^2\p$ time; 
                $\n^2\o^2 + \n\o\left[\y^\li + \p^\ti\right] + \n\y + \p$ memory.
            \end{tcolorbox}
            In summary, \jc{} performs $\n\o$ forward passes followed by an expensive $\n^2\o^2\p$ contraction. Next we demonstrate how to reduce the contraction cost.

        \subsection{\texorpdfstring{\ntvp}{NTK-vector Products}}\label{sec:implicit} 

            Consider the NTK-vector product function (for $\n = 1$):
            $$\Theta_\theta^f\textrm{VP} \colon v \in\mathbb{R}^\o\mapsto\Theta_\theta^f v\in \mathbb{R}^\o.$$
            Taking the NTK-vector product with $\o$ columns of the identity matrix $I_\o$ yields the full NTK, i.e.
            $\Theta_\theta^f I_\o = \Theta_\theta^f.$ 
            Expanding $\Theta_\theta^f\textrm{VP}(v)$ as 
            \begin{align}\label{eq:ntk_vp_main}
                \Theta_\theta^f v 
                &= 
                \left[{\partial f\left(\theta, x_1\right)}\big/{\partial\theta}\right]\left[{\partial f\left(\theta, x_2\right)}\big/{\partial\theta}\right]^T v =\\
                &=\left[{\partial f\left(\theta, x_1\right)}\big/{\partial\theta}\right]\textrm{VJP}^f_{\left(\theta, x_2\right)}\left(v\right) =\\ 
                &=\textrm{JVP}^f_{\left(\theta, x_1\right)}\left[\textrm{VJP}^f_{\left( \theta, x_2\right)}\left(v\right)\right],
            \end{align}
            where we have observed that the NTK-vector product can be expressed as a composition of a JVP and a VJP. The cost of computing $\Theta_\theta^f$ is then asymptotically the cost of the \jac, since it consists of $\o$ VJPs followed by $\o$ (cheaper) JVPs, therefore $\o\left[\fp\right]$ time and $\o\left[\y^\li + \p^\ti\right] + \y + \p$ memory. In the batched setting \cref{eq:ntk_vp_main} is repeated for each pair of inputs, and therefore time increases by a factor of $\n^2$ to become $\n^2\o\left[\fp\right]$. However, the memory cost grows only linearly in $\n$ (except for the cost of storing the NTK of size $\n^2\o^2$), since intermediate primitive outputs and tangents/cotangents can be computed for each batch $x_1$ and $x_2$ separately and then reused for every pairwise combination. Therefore the memory cost is asymptotically the cost to store the NTK and compute the \jac. Altogether,
            \begin{tcolorbox}
                \ntvp~cost 
                $\n^2\o\left[\fp\right]$ time; 
                $\n^2\o^2 + \n\o\left[\y^\li + \p^\ti\right] + \n\y + \p$ memory.
            \end{tcolorbox}
            
            In summary, \ntvp{} eliminate the costly $\n^2\o^2\p$ contraction of \jc, but perform $\n^2\o$ forward passes (as opposed to $\n\o$), and the memory requirement is identical. As a result, this method is beneficial for small $\n$, and for networks with a cheap forward pass \fp{} relative to $\o\p$, which is always the case for FCNs (\sref{sec:example_fcn}), but not necessarily for CNNs (\sref{sec:example_cnn_main_text}).

        \subsection{\texorpdfstring{\sd}{Structured Derivatives}}\label{sec:str_derivatives}
        
            Rewriting $\Theta_\theta^f$ from \cref{eq:ntk_def_main} using the chain rule in terms of the primitive outputs $y^\li$, we find:
            \begin{align}\label{eq:expanded_ntk_main}
                \Theta_\theta^f =
                \sum_{\ti,\li_1,\li_2}&\left(\frac{\partial f_1}{\partial y_1^{\li_1}}\frac{\partial y_1^{\li_1}}{\partial \theta^{\ti}}\right)\left(\frac{\partial f_2}{\partial y_2^{\li_2}}\frac{\partial y_2^{\li_2}}{\partial \theta^{\ti}}\right)^T 
                \\ =
                \sum_{\ti,\li_1,\li_2}
                &\left({\frac{\partial f_1}{\partial y_1^{\li_1}}\frac{\partial y_1^{\li_1}}{\partial \theta^{\ti}}\frac{\partial y_2^{\li_2}}{\partial \theta^{\ti}}^T\frac{\partial f_2}{\partial y_2^{\li_2}}^T}\right) \\ \eqqcolon \sum_{\ti,\li_1,\li_2}&\Theta_\theta^f\left[\ti, \li_1,\li_2\right],
            \end{align}
            where we define $f_i \coloneqq f(\theta, x_i)$, and only consider $\partial y_i^{\li_i} / \partial \theta^\ti$ to be non-zero if $\theta^\ti$ is a direct input to $y_i^{\li_i}$. We have also defined $\Theta_\theta^f\left[\ti, \li_1,\li_2\right]$ to be individual summands.
            
            Both \jc~and \ntvp~perform this sum of contractions via VJPs and JVPs, without explicit instantiation of primitive Jacobians $\partial y^{\li_i}_i/\partial \theta^\ti$. However, while VJPs and JVPs themselves are guaranteed to be computationally optimal (\sref{sec:jvp_vjp_costs}), higher order computations like their composition (\ntvp) or contraction (\jc) are not. Specifically, each $\Theta_\theta^f\left[\ti, \li_1,\li_2\right]$ from \cref{eq:expanded_ntk_main} is a matrix-Jacobian-Jacobian-matrix product (\textbf{MJJMP}), which, as we will show shortly, can't always be evaluated optimally with 
            VJPs and JVPs.
            
            The idea of \sd~is to design rules for efficient computation of MJJMPs, similarly to AD rules for JVPs and VJPs.  
            
            From~\cref{eq:expanded_ntk_main}, in the general case this requires hand-made rules for all pairwise combinations of primitives $y_1$ and $y_2$, of which there are $136^2 > 10,000$ in JAX, and even more in Tensorflow \citep{abadi2016tensorflow} and PyTorch \citep{pytorch} (see \sref{sec:jax}). We dramatically reduce this number by:
            
            \textbf{1. Linearization.} 
                It follows from \cref{eq:jvp_vjp_main}, that $\Theta_\theta^f = \Theta_\theta^{\textrm{JVP}_{\left(\theta, \cdot\right)}^f}$, i.e. the NTK of $f$ evaluated at parameters $\theta$ is equal to the NTK of the JVP of $f$ given primal $\theta$. JVP is a linear function of input tangents $\theta$, and therefore we only need to implement efficient MJJMPs for linear primitives, of which JAX has only 56.\footnote{A linear function can contain nonlinear primitives. However, linearizing any function in JAX is guaranteed to produce only linear primitives \citep{frostig2021decomposing, radul2022you}.}
                
            \textbf{2. MJJMPs through structured derivatives.} 
                We further reduce the necessary MJJMP rule count from $56^2$ down to only $56$ by decomposing an MJJMP rule into two parts:
                \begin{enumerate}
                    
                    \item \textbf{Structured derivative rule}. Given a single primitive $y$, this rule identifies the smallest subarray of $\partial y/\partial \theta^\ti$ sufficient to reconstruct the entire primitive Jacobian $\partial y/\partial \theta^\ti$, and the (constant and negligible in memory size) metadata necessary for the reconstruction. For example, if $x\in\mathbb{R}^\w$, $\theta^\ti\in\mathbb{R}^{\w\times\w}$, and  $y\left(\theta^\ti\right) = \theta^\ti x \in\mathbb{R}^{\w}\,$ (matrix-vector multiplication), then $\partial y/\partial \theta^\ti = I_{\w} \otimes x^T\in\mathbb{R}^{\w\times\w^2}$, and the rule will indicate that (1) only the subarray $\left[\partial y/\partial \theta^{\ti}\right]_{1,:\w}\in\mathbb{R}^{1\times\w}$,\footnote{We define $\left[A\right]_{i, :j}\coloneqq\left[A_{i, 1},\dots,A_{i, j}\right]\in\mathbb{R}^{1\times j}$.} needs to be computed (which is equal to $x^T$ in this case), and (2) that the entire primitive Jacobian can be reconstructed as $\partial y/\partial \theta^\ti = I \otimes \left[\partial y/\partial \theta^{\ti}\right]_{1, :\w}$. In other words, this rule annotates linear primitives $y$ with the structure of their Jacobians, such as block diagonal, constant-block diagonal, or tiling along certain dimensions.
                    
                    \item \textbf{MJJMPs with structured Jacobians.} Given input tensors $A$, $B$, $C$, $D$, where $B$ and $C$ are provided in the structured form as described above (i.e. only small subarrays along with their metadata) this rule efficiently computes the 4-way contraction $A B C D$ (i.e. the NTK summand $\Theta_\theta^f\left[\ti, \li_1,\li_2\right]$). This amounts to using \code{np.einsum} with the optimal contraction order and adjusting its instructions based on provided metadata. For example, if $B = I_{\w} \otimes b^T\in\mathbb{R}^{\w\times \w^2}$ and $C = I_{\w} \otimes c^T\in\mathbb{R}^{\w\times \w^2}$ (for $b, c \in \mathbb{R}^{\w}$), then
                    \begin{align}\label{eq:mixed_product_example}
                        A B C D &= 
                        A\left(I \otimes b^T\right)\left(I \otimes c^T\right)^T D = \\ &=
                        A\left(I \otimes b^T c\right) D = 
                        \left(b^T c\right) A D,
                    \end{align}
                    where were able to pull out $b^T c$ since it is a scalar. As we will see in \sref{sec:examples} and \sref{sec:more_structure}, this and other similar contraction rules can enable significant speedups.
                
                \end{enumerate}
            
                Therefore we avoid implementing $56^2$ MJJMP rules by instead having (1) a single routine to perform 4-way tensor contractions with structured tensors, and (2) $56$ rules annotating the structure in the $56$ linear primitive Jacobians. We list all these structures and associated MJJMP costs in \sref{sec:more_structure}. Our approach does not guarantee optimality for the NTK of an arbitrary function, however, as we show in \sref{sec:examples}, it is asymptotically better than \jc{} for FCNs and CNNs, and can provide orders of magnitude speedups in much more complex contemporary ImageNet models (\sref{sec:experiments}).
            
            \textbf{3. Focusing on MJJMPs for typical operations.} 
                Many of the $56$ linear JAX primitives are trivial to implement or rarely arise in NNs. At the time of writing we have only annotated  $21$ linear primitives (\cref{tab:list_of_primitives}), which was  sufficient for the empirical speedups observed in \sref{sec:examples} and \sref{sec:experiments}.

            \textbf{Summary.} 
                \sd{} amount to evaluating the sum of MJJMPs in \cref{eq:expanded_ntk_main}, where (1) only small subarrays of primitive Jacobians $\partial y_i^{\li_i} / \partial \theta^\ti$ are instantiated, and (2) MJJMPs leverage the structure of these primitive Jacobians for efficient contractions. Together, this incurs
                \begin{enumerate}
                     
                     \item The cost of computing primitive output cotangents $\partial f_i / \partial y_i^{\li_i}$ for \cref{eq:expanded_ntk_main}, which is equivalent to the cost of the reverse mode \jac{} (\sref{sec:jacobian}), \emph{less} the cost of computing ($\n\o\p$) and storing ($\n\o\p^\ti$) weight-space cotangents $\partial f_i/\partial \theta^\ti$, since they aren't used in \cref{eq:expanded_ntk_main}, i.e. $\n\o\left[\fp\right]$ time\footnote{Note that $\n\o\p$ time saving is not reflected in the asymptotic cost since it is dominated by $\n\o\left[\fp\right]$ required to compute primitive output cotangents. However, as we will see in \cref{fig:fcn_flops}, it often provides a substantial practical benefit, up to allowing to compute the NTK faster than computing the two Jacobians themselves.} and $\n\o\y^\li + \n\y + \p$ memory.
                     
                     \item The cost of computing primitive Jacobian $\partial y_i^{\li_i} / \partial \theta^\ti$ subarrays, denoted as $\j^{\li_i}_\ti$ with $\j \coloneqq \sum_{\ti,\li_1} \j^{\li_1}_\ti + \sum_{\ti,\li_2} \j^{\li_2}_\ti$. This amounts to $\n\j$ time and $\n\j^{\li}_\ti$ memory.
                     
                     \item The cost of evaluating $\Theta_\theta^f\left[\ti, \li_1,\li_2\right]$ via the efficient MJJMP, which we denote as 
                     simply \textbf{MJJMP} and substitute specific values based on the primitive. Required memory to store the result is $\n^2\o^2$.
                 
                \end{enumerate}
            
            We add up these costs below and in \cref{tab:intro_summary_informal}, and show in \sref{sec:examples} and \sref{sec:experiments} how they are beneficial in most practical settings. 
            \begin{tcolorbox}
                \sd{} cost 
                $\n\o\left[\fp\right] + \textbf{MJJMP} + \n\left[\j - \o\p\right]$ time; 
                $\n^2\o^2 + \n\o\y^\li + \n\j^{\li}_{\ti} + \n\y + \p$ memory.
            \end{tcolorbox}

\setlength\arrayrulewidth{0.6pt}

\begin{table*}[!htbp]
    \resizebox{\textwidth}{!}{
        \begin{tabular}{|l|l|l|l|}
            \hline
            &&&\\[-1.05em]
            Method                 & Time                        & Memory & Use when                                 
            \\ 
            \hline
            &&&\\[-1.05em]
            \jc & $\n\hphantom{^2}\o \left[\textrm{\fp}\right] + \n^2\o^2 \p$            & $\n^2\o^2 + \n\o\left[\y^\li + \p^\ti\right] + \n\y + \p$ & $\p < \y$, small $\o$ 
            \\ 
            \hline
            &\cellcolor{yellow!10}&&\\[-1.05em]
            \ntvp & \cellcolor{yellow!10}$\n^2\o\left[\textrm{\fp}\right]$ & $\n^2\o^2 + \n\o\left[\y^\li+ \p^\ti\right] + \n\y + \p$ & $\fp < \o\p$, large $\o$, small $\n$
            \\ 
            \hline
            &\cellcolor{yellow!10}&\cellcolor{yellow!10}&\\[-1.05em]
            \sd & \cellcolor{yellow!10}$\n\hphantom{^2}\o\left[\fp\right] + \textbf{MJJMP} + \n\j$ & \cellcolor{yellow!10}$\n^2\o^2 + \n\o\y^\li + \n\j^{\li}_{ \ti}\, + \n\y + \p $ & $\fp > \o\p$, large $\o$, large $\n$
            \\ 
            \hline
            \end{tabular}
            }
            \caption{\textbf{Generic NTK computation costs}. \ntvp~trade-off contractions for more \fp. \sd~usually save both time and memory. See \sref{sec:notation} and \sref{sec:str_derivatives} for notation, and \sref{sec:glossary} for a glossary of symbols.
            }\label{tab:intro_summary_informal}\vspace{0.25cm}
            \resizebox{\textwidth}{!}{\begin{tabular}{|l|l|l|l|}
            \hline
            &&&\\[-1.05em]
            Method                 & Time                        & Memory & Use when                                 
            \\ 
            \hline
            &&&\\[-1.05em]
            \jc & $\hphantom{\n^2\o\t\w^2 + \,\,\,} \n^2\o^2\t\w^2 + \n^2\o^3\w$            & $\n^2\o^2 + \n\o\w^2 + \n\o^2\w + \n\t\w + \t\w^2$ & Don't 
            \\ 
            \hline
            &\cellcolor{green!10}&&\\[-1.05em]
            \ntvp & \cellcolor{green!10}$\n^2\o\t\w^2 + \n^2\o^2\hphantom{\t}\w\hphantom{^2}$ & $\n^2\o^2 + \n\o\w^2 + \n\o^2\w + \n\t\w + \t\w^2$ & $\o>\w$ or $\n = 1$
            \\ 
            \hline
            &\cellcolor{green!10}&\cellcolor{green!10}&\\[-1.05em]
            \sd~& \cellcolor{green!10}$\n\hphantom{^2}\o\t\w^2 + \n^2\o^2\t\w\hphantom{^2} + \n^2\o^3$ & \cellcolor{green!10}$\n^2\o^2 + \n\o\w\hphantom{^2 + \n\o^2\w\Bigg|} + \n\t\w + \t\w^2$ & $\o<\w$ or $\t = 1$
            \\
            \hline
        \end{tabular}
    }
    \caption{
        \textbf{FCN NTK computation cost.} The costs are obtained by substituting into \cref{tab:intro_summary_informal} specific values for $\fp$, $\p$, $\y$, $\j$, and \textbf{MJJMP} that correspond to an FCN as described in \sref{sec:example_fcn}. \ntvp~allow a reduction of the time complexity, while \sd~reduce both time and memory complexity. See \cref{fig:fcn_flops} for empirical confirmation with FLOPs and wall-clock time. See \sref{sec:example_fcn} for discussion and notation (\sref{sec:glossary} for the full glossary of symbols).
    }\label{tab:intro_summary}
    
    \vspace{0.25cm}
    \setlength\arrayrulewidth{0.8pt}
    
    \resizebox{\textwidth}{!}{
        \begin{tabular}{|l|l|l|l|}
            \hline
            &&&\\[-0.7em]
            Method                 & Time                        & Memory   & Use when                          
            \\[4pt]
            \hline
            &&&\\[-0.7em]
            \jc & $\n\hphantom{^2}\o\left[\t\d\f\w^2 + \o\w\right] + \n^2\o^2\left[\t\f\w^2 + \o\w\right]$ & $\n^2\o^2 + \n\o\left[\d\w + \f\w^2 + \o\w\right] + \n\left[\t\d\w \right] + \left[\t\f\w^2 + \o\w^2\right]$ & $\d > \o\w$ 
            \\[4pt]
            \hline
            &\cellcolor{yellow!10}&&\\[-0.7em]
            \ntvp & \cellcolor{yellow!10}$\n^2\o\left[\t\d\f\w^2 + \o\w\right]$ & $\n^2\o^2 + \n\o\left[\d\w + \f\w^2 + \o\w\right] + \n\left[\t\d\w\right] + \left[\t\f\w^2 + \o\w^2\right]$& $\n = 1$
            \\[4pt]
            \hline
            &\cellcolor{green!10}&\cellcolor{yellow!10}&\\[-0.7em]
            \sd & \cellcolor{green!10}$\n\hphantom{^2}\o\left[\t\d\f\w^2 + \o\w\right] + \n^2\o^2\left[\t\min\left(\f\w^2, \d\w + \frac{\d\f\w^2}{\o}, \d\w + \frac{\d^2\w}{\o} + \frac{\d^2\f\w}{\o^2}\right) + \o\right]$ & \cellcolor{yellow!10}$\n^2\o^2 + \n\o\left[\d\w\right] \quad + \quad \n\d\f\w\,\,+ \n\left[\t\d\w\right] + \left[\t\f\w^2 + \o\w^2\right]$ & $\d < \o\w$
            \\[6pt]
            \hline
        \end{tabular}
    }
    \caption{
        \textbf{CNN NTK computation cost} for a CNN with $\d$ pixels and filter size $\f$. \sd~reduce time complexity, and have lower memory cost if $\d < \o\w$, which is a common setting. See \cref{fig:resnet_secs} for experiments with ResNets, \sref{sec:example_cnn_main_text} for discussion, \cref{tab:intro_summary} for FCN, \cref{tab:intro_summary_informal} for generic cost analysis, and \sref{sec:glossary} for a glossary of symbols.
    }\label{tab:intro_summary_1x1}\vspace{-0.06cm}
\end{table*}

\section{Examples}\label{sec:examples}

    The three algorithms in \sref{sec:main_text} (and our implementations) apply to any differentiable function $f$, but the resulting complexities depend on variables such as $\fp$ and $\p$, which depend on $f$  (\cref{tab:intro_summary_informal}). Below we compute and compare all three complexities for a deep FCN (\sref{sec:example_fcn}) and CNN (\sref{sec:example_cnn_main_text}), summarizing the results in \cref{tab:intro_summary} and \cref{tab:intro_summary_1x1} respectively.

    \subsection{FCN NTK Complexity}\label{sec:example_fcn}
    
        We apply our algorithms from \sref{sec:main_text} to FCNs with $\t$ hidden layers of width $\w$. For simplicity we assume no biases, and $x \in\mathbb{R}^\w$, i.e. inputs of same size as the width. We define $y^\ti \coloneqq \theta^\ti x^{\ti}$, and $x^\ti \coloneqq \phi\left(y^{\ti-1}\right)$ for $\ti > 0$, with $x^0 \coloneqq x$. Output is  $f\left(x,\theta\right) \coloneqq y^\t$. See \cref{fig:fcn_legend} (top) for $\t = 2$.
        
        In this case $\l = 2\t + 1$ ($\t + 1$ matrix-vector multiplications and $\t$ nonlinearities), $\p^\ti = \w^2$ for $\ti < \t$ and $\o\w$ for the top layer $\ti = \t$, $\p = \t\w^2 + \o\w$, $\y^\li = \w$ for $\li < \l$ and $\o$ for $\li = \l$, and $\y \sim \t\w + \o$. Finally, a single forward pass $\fp \sim \t\w^2 + \o\w$ time and $\w^2 + \o\w$ memory.
        
        Plugging the above into the cost of the baseline algorithm \jc{} in \sref{sec:vanilla}, we obtain
        \begin{tcolorbox}
            FCN \jc{} costs 
            $\n^2\o^2\t\w^2 + \n^2\o^3\w$ time; 
            $\n^2\o^2+\n\o\w^2+\n\o^2\w + \n\t\w+\t\w^2$ memory.
        \end{tcolorbox}
        Similarly, the cost of \ntvp{} from \sref{sec:implicit} is
        \begin{tcolorbox}
            FCN \ntvp{} cost 
            $\n^2\o\t\w^2 + \n^2\o^2\w$ time; 
            $\n^2\o^2+\n\o\w^2+\n\o^2\w + \n\t\w+\t\w^2$ memory.
        \end{tcolorbox}
        For \sd{} (\sref{sec:str_derivatives}), we additionally need to derive values of $\j$ and \textbf{MJJMP}. For an arbitrary primitive, $\j$ and \textbf{MJJMP} costs are derived by (1) looking up the type of structure in the primitive Jacobian in \cref{tab:list_of_primitives}, followed by (2) extracting the costs for a given structure from \cref{tab:complexities_structures} (see \sref{sec:more_structure}). We apply this formal approach in \sref{sec:example_fcn_mjjmp}, but for demonstration purposes below present the derivation that does not require referencing the \hyperref[sec:appendix]{Appendix}.
        
        We first note that we only need to consider $\li_1 = \li_2 = 2\ti + 1$ indices in \cref{eq:expanded_ntk_main}, since all other summands are zero due to absence of weight sharing between layers. For matrix-vector multiplication $y_i^{\ti} = \theta^{\ti}x_i^{\ti}$ our rules indicate (per example given in \sref{sec:str_derivatives}) that $\partial y_i^{\ti}/\partial \theta^\ti = I_{\w} \otimes \left[\partial y_i^{\ti}/\partial \theta^\ti\right]_{1,:\w}\in\mathbb{R}^{\w\times\w^2}$, and command to only compute $\left[\partial y_i^{\ti}/\partial \theta^\ti\right]_{1,:\w}\in\mathbb{R}^{1\times \w}$ (which is ${x_i^{\ti}}^T$ in this case). Therefore $\j_{\ti}^{2\ti + 1} = \w$, and $\j = 2\sum_{\ti=1}^{\t} \j^{2\ti + 1}_{\ti} \sim \t\w$. 
    
        Finally, the efficient MJJMP for this structured $\partial y_i^{\ti}/\partial \theta^\ti$ can be computed, analogously to \cref{eq:mixed_product_example}, as follows for $\ti < \t$:
        \begin{align}
            \underbrace{\Theta_\theta^f\left[\ti, 2\ti + 1, 2\ti + 1\right]}_{\o\times\o} = 
            \frac{\partial f_1}{\partial y_1^{\ti}}\frac{\partial y^{\ti}_1}{\partial \theta^\ti}{\frac{\partial y^{\ti}_2}{\partial \theta^\ti}}^T\frac{\partial f_2}{\partial y_2^{\ti}}^T
            = \\ =
            \underbrace{\frac{\partial f_1}{\partial y_1^{\ti}}}_{\o\times\w}
            \left(I_{\w}\otimes \underbrace{{x^{\ti}_1}^T}_{1\times\w}\right)\left(I_{\w}\otimes \underbrace{{x^{\ti}_2}^T}_{1\times\w}\right)^T\underbrace{\frac{\partial f_2}{\partial y_2^{\ti}}^T}_{\w\times\o} = \\ =
            \left(\underbrace{{x_1^{\ti}}^T}_{1\times\w}\underbrace{{x_2^{\ti}}}_{\w\times 1}\right)\underbrace{\frac{\partial f_1}{\partial y_1^{\ti}}}_{\o\times\w}\underbrace{{\frac{\partial f_2}{\partial y_2^{\ti}}}^T}_{\w\times\o},
        \end{align}
        which can be contracted in only $\o^2\w$ time. An analogous derivation applied to $\ti = \t$ yields $\o^3 + \w$ time. Therefore the total contraction cost is $\textbf{MJJMP} \sim \n^2\t\o^2\w + \n^2\o^3$, when accounting for depth $\t$ and batch size $\n$. Altogether,
        \begin{tcolorbox}
            FCN \sd{} cost 
            $\n\o\t\w^2 + \n^2\o^2\t\w + \n^2\o^3$ time; 
            $\n^2\o^2+\n\o\w+\n\t\w+\t\w^2$ memory.
        \end{tcolorbox}
    
        \textbf{Summary.} 
            We summarize all FCN costs in \cref{tab:intro_summary}. We conclude that \sd~and \ntvp~allow a reduction in the time cost of NTK computation in different ways, while \sd~also reduce memory requirements. \sd~are beneficial for wide networks, with large $\w$, and \ntvp~are beneficial for networks with large outputs $\o$. 
        
\begin{figure*}
            \centering
            \hfil\textbf{FLOPs (per NTK entry)}\hfil\hfil\hfil\textbf{Wall-clock time (TPUv3)}\hfil\\
            \includegraphics[width=0.495\textwidth]{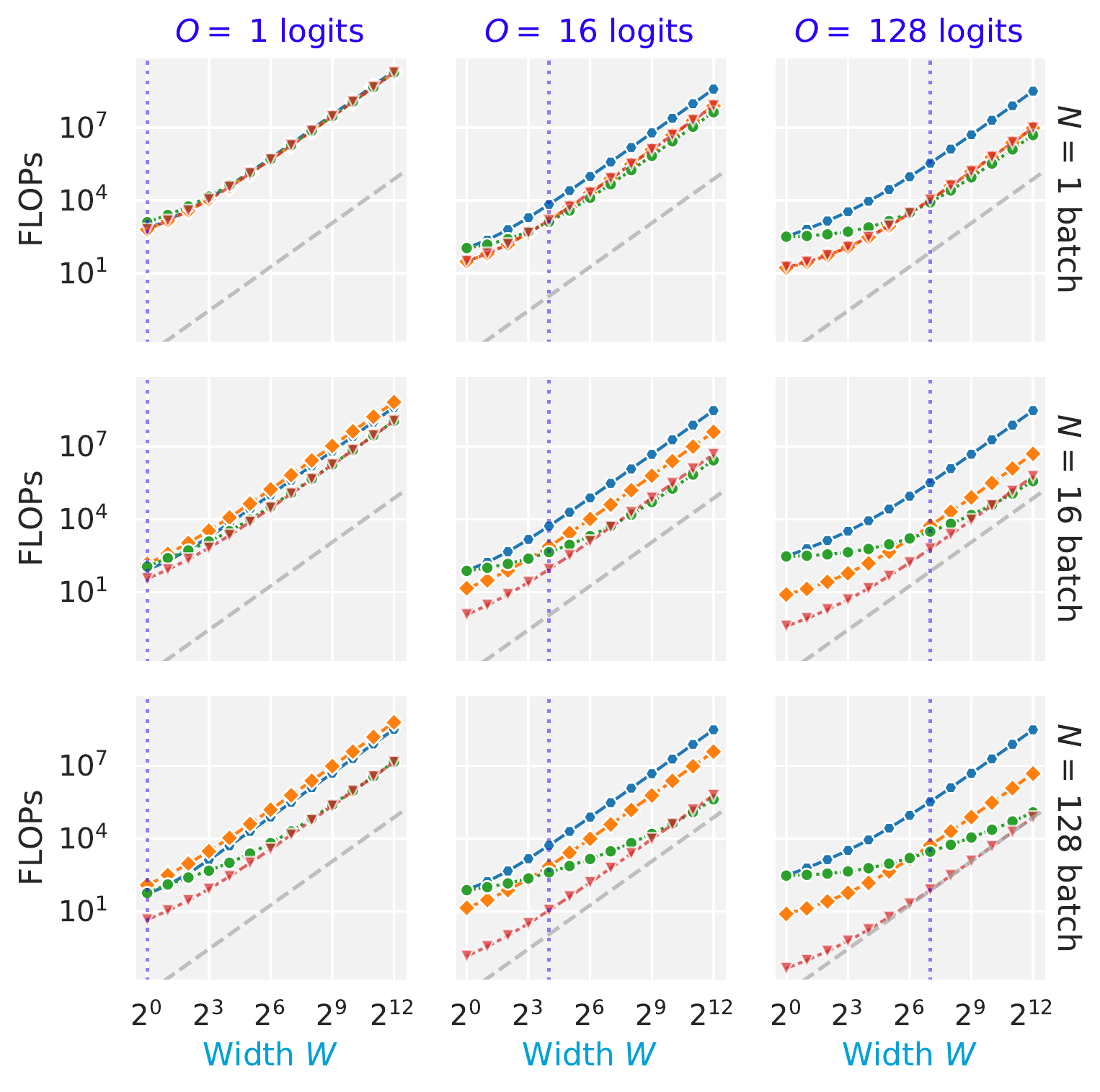}
            \includegraphics[width=0.495\textwidth]{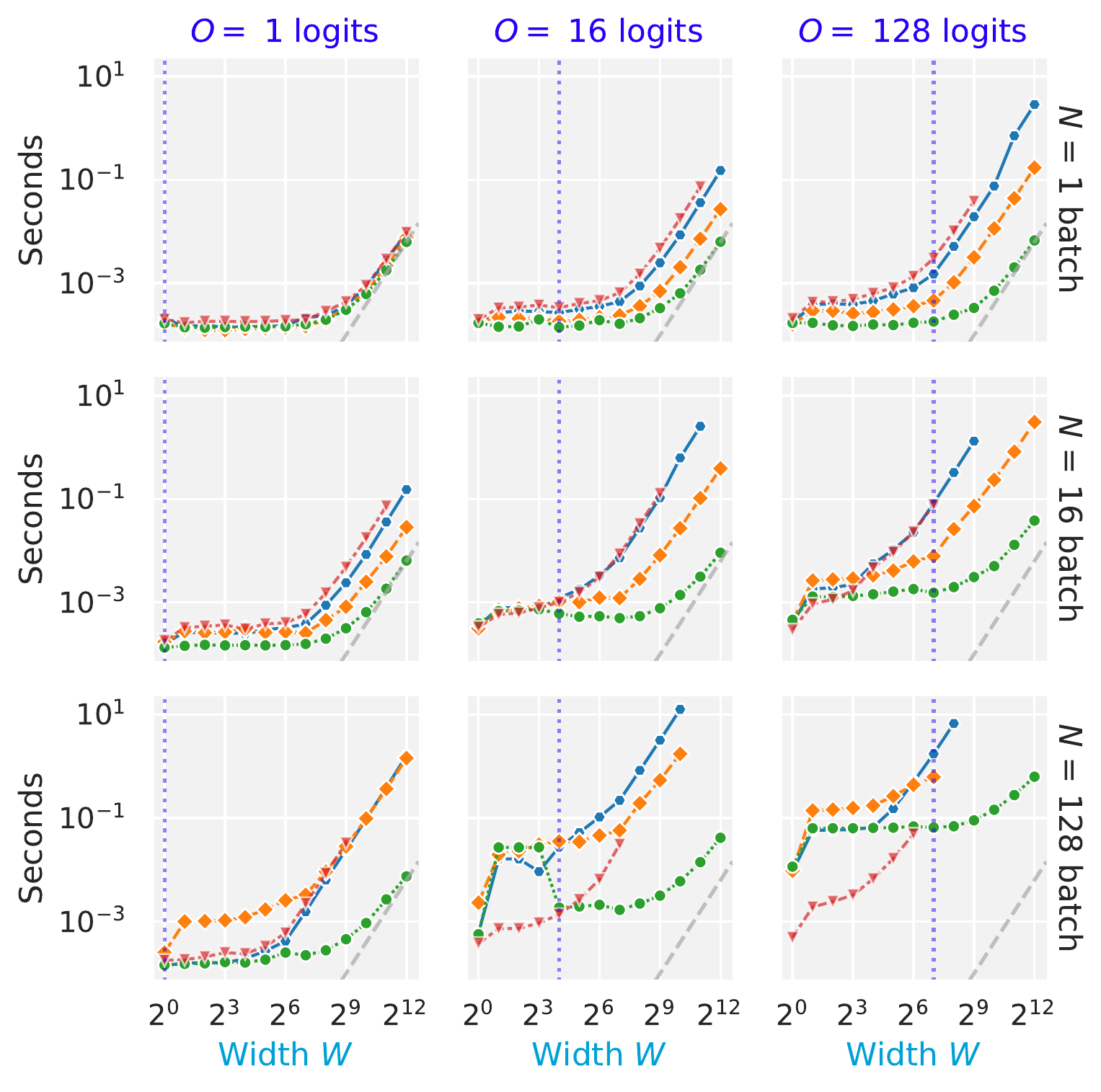}
            
            \vspace{0.1cm}
            
            \hphantom{123\,\,}\includegraphics[width=0.92\textwidth]{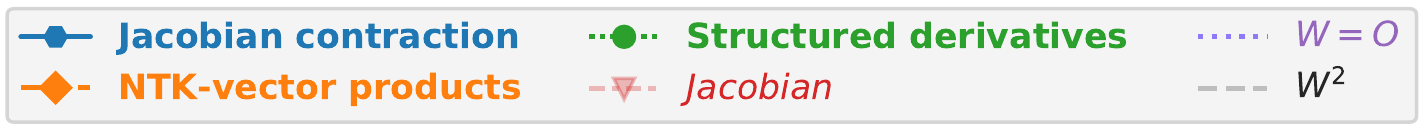}
            \vspace{0.3cm}
            \caption[]{
                \textbf{FLOPs (left) and wall-clock time (right) of computing the NTK for a 10-layer ReLU FCN.} As predicted by \cref{tab:intro_summary}, our methods almost always outperform \jc, allowing orders of magnitude speed-ups and memory improvements for realistic problem sizes.  
                \vspace{0.2cm}
                
                \begin{tabular}{p{0.47\linewidth} p{0.47\linewidth}}
                    \textbf{Left: FLOPs per NTK entry.} We confirm several specific theoretical predictions from \sref{sec:example_fcn}:
                    \begin{enumerate}
                        \item \ntvp~are the best performing method for $\n = 1$, and have cost equivalent to \jac~for any width $\w$ or output size $\o$ (top row); 
                        \item \ntvp~offer an $\o$-fold improvement over \jc~(left to right columns); 
                        \item \ntvp~are equivalent to \jc~for $\o = 1$ (leftmost column);
                        \item \sd~outperform \ntvp~$\operatorname{iff}$ $\o<\w$ ($\o = \w$ are plotted as pale vertical lines, which is where \sd~and \ntvp~intersect);
                        \item \sd~approach the cost of \jac~in the limit of large width $\w$ (left to right);
                        \item All methods, as expected, scale quadratically with width $\w$ (pale grey dashed line depicts $\w^2$ scaling).
                    \end{enumerate}
                    &
                    \textbf{Right: Wall-clock runtime.} In real applications, given the \xla~compiler and hardware specifics, we observe that: 
                    \begin{enumerate}
                        \item \ntvp~improve upon \jc~for $\o > 1$, but the effect is not perfectly robust (see bottom row for small $\w$ and \cref{fig:fcn_secs_other}, notably GPU platforms);
                        \item \sd~robustly outperform all other methods, including simply computing the \jac, as discussed in \sref{sec:str_derivatives};
                        \item \sd~have lower memory footprint, and reach up to 8x larger widths (bottom right; missing points indicate out-of-memory), i.e. can process models up to 64x larger than other methods, as discussed in \sref{sec:str_derivatives};
                        \item All methods have a smaller memory footprint than \jac~(see \sref{sec:jacobian}).
                    \end{enumerate}
                \end{tabular}
                \textbf{More:} see \cref{fig:fcn_secs_other} for other hardware platforms, and \sref{sec:experimental_definition} for details.
            }\label{fig:fcn_flops}
        \end{figure*}

            We confirm our predictions with FLOPs measurements in \cref{fig:fcn_flops}. We further confirm our methods provide orders of magnitude speed-ups and memory savings on all major hardware platforms in \cref{fig:fcn_flops} (right) and \cref{fig:fcn_secs_other}. However, we notice that time measurements often deviate from predictions due to unaccounted constant overheads of various methods, hardware specifics, padding, and the behavior of the \xla~compiler. We find \sd~to almost always outperform \ntvp.

    \begin{figure*}
        \centering
        \includegraphics[width=\textwidth]{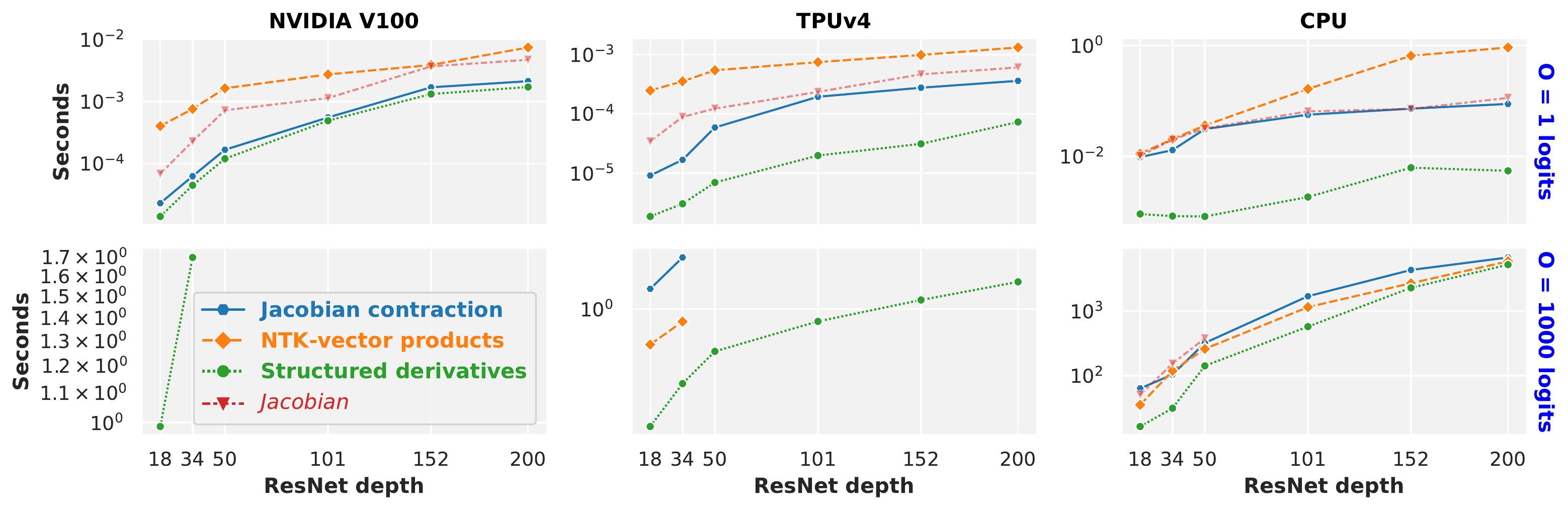}
        \caption{
            \textbf{Wall-clock time of computing an NTK for several ResNet sizes on a pair of ImageNet images}. \sd~allow the NTK to be computed faster and for larger models (see bottom row -- missing points indicate out-of-memory). \ntvp, as predicted in \sref{sec:implicit} and \cref{tab:intro_summary_informal}, are advantageous for large $\o$ (bottom row), but are suboptimal when the cost of the forward pass $\fp$ is large relative to the output size and the number of parameters $\o\p$, e.g. when there is a lot of weight sharing (see \cref{tab:intro_summary_1x1} and \cref{tab:intro_summary_informal}), which is the case for convolutions, notably for $\o = 1$ (top). See \cref{fig:imagenet_other_secs} for more ImageNet models, \sref{sec:example_cnn_main_text} for CNN NTK complexity analysis, and \sref{sec:experimental_definition} for experimental details.
        }\label{fig:resnet_secs}
        \vspace{-0.1cm}
    \end{figure*}

    \subsection{CNN NTK Complexity}\label{sec:example_cnn_main_text}
        
        We perform analogous derivations for a CNN with $\w$ channels, $\d$ pixels, and filter size $\f$ in \sref{sec:example_cnn}. We arrive at \cref{tab:intro_summary_1x1}, and make two observations specific to CNNs. 
        
        First, the speeds of \ntvp{} and \jc{} are now much more similar, due to the higher cost of the forward pass \fp{} relative to $\p$ (i.e. weight sharing), and how they perform will depend on the specific values of parameters. We confirm this in our experiments on  ImageNet models in \sref{sec:experiments}, where \ntvp{} typically underperform for $\o = 1$, but outperform for $\o = 1000$.
        
        Secondly, \sd{} continue to perform faster than \jc{}, but the relative memory costs depend on other hyperparameters, requiring $\d < \o\w$. This is a common case for ImageNet with $\o = 1000$, and is confirmed in our experiments in  \cref{fig:resnet_secs} and \cref{fig:imagenet_other_secs} (bottom).

\section{ImageNet Experiments}\label{sec:experiments}
    
    In \sref{sec:examples} we have derived asymptotic time and memory benefits of \ntvp{} and \sd{} over the baseline \jc{} for FCNs and CNNs. However, contemporary architectures rarely resemble vanilla feedforward networks, but instead result in  much more complex computational graphs comprised of many different primitives, making complexity analysis impractical. 

    We therefore evaluate our methods in the wild, and confirm computational benefits on full ImageNet models in \cref{fig:resnet_secs} (ResNets, \citet{he2016deep}) and \cref{fig:imagenet_other_secs} (WideResNets, \citet{zagoruyko2016wide}; Vision Transformers and  Transformer-ResNet hybrids \citet{dosovitskiy2021an, steiner2021augreg}; and MLP-Mixers \citet{tolstikhin2021mlpmixer}). Computing the full $\o\times\o=1000\times 1000$ NTK is often only possible with \sd.

\section{Implementation}\label{sec:implementation}
    
    All algorithms are implemented in JAX\footnote{See \sref{sec:jax} for discussion about other frameworks.}
    \citep{jax2018github} and integrated into Neural Tangents \citep{neuraltangents2020}. \jc~and \ntvp~are built with core operations such as \code{vjp}, \code{jvp}, and \code{vmap}. \sd~are implemented as a \href{https://jax.readthedocs.io/en/latest/notebooks/Writing_custom_interpreters_in_Jax.html}{Jaxpr interpreter}, built on top of the JAX reverse mode AD interpreter. 
    
    Owing to the nuanced trade-offs between different methods in the general case, we release all implementations within a single function that allows the user to manually select implementation. We also include an automated setting which will perform FLOPs analysis for each method at compilation time and automatically choose the most efficient one.

\section{Conclusion}\label{sec:conclusion}
    
    We have performed the first extensive analysis of the computational complexity of the NTK, and have shown how it can be improved dramatically with mixed-order AD (\ntvp), or with a custom interpreter for more efficient higher-order AD operations (\sd).
    
    The NTK computation is similar to many other objects of interest in machine learning, such as the Gauss-Newton or the Fisher Information matrix, and we look forward to extensions of our algorithms to more settings in future work.

\pdfbookmark{Acknowledgements}{unnumbered}
\section*{Acknowledgements}
    We thank Lechao Xiao for useful discussion, review and comments on the initial version of this manuscript, and Jaehoon Lee for useful discussion and code review.
    
    We also thank Shaobo Hou for his work on and help with TF2Jax, and the JAX team for their help and advice on JAX and Jax2TF.
\setcounter{section}{0}

\pdfbookmark{References}{unnumbered}
\bibliography{main}
\bibliographystyle{icml2022}
\setcounter{section}{0}

\appendix
\onecolumn
\appendix
\pdfbookmark{Appendix}{unnumbered}
\section*{Appendix}\label{sec:appendix}
\setcounter{section}{0}
\renewcommand{\thesubsection}{\Alph{subsection}}

    \subsection{Additional Figures}\label{sec:additional_figures}

        \begin{figure}[!hb]
            \centering
            \hfil\hfil\textbf{CPU (Skylake)}\hfil\hfil\hfil\hfil\textbf{NVIDIA V100}\hfil\hfil\hfil\\
            \includegraphics[width=0.48\textwidth]{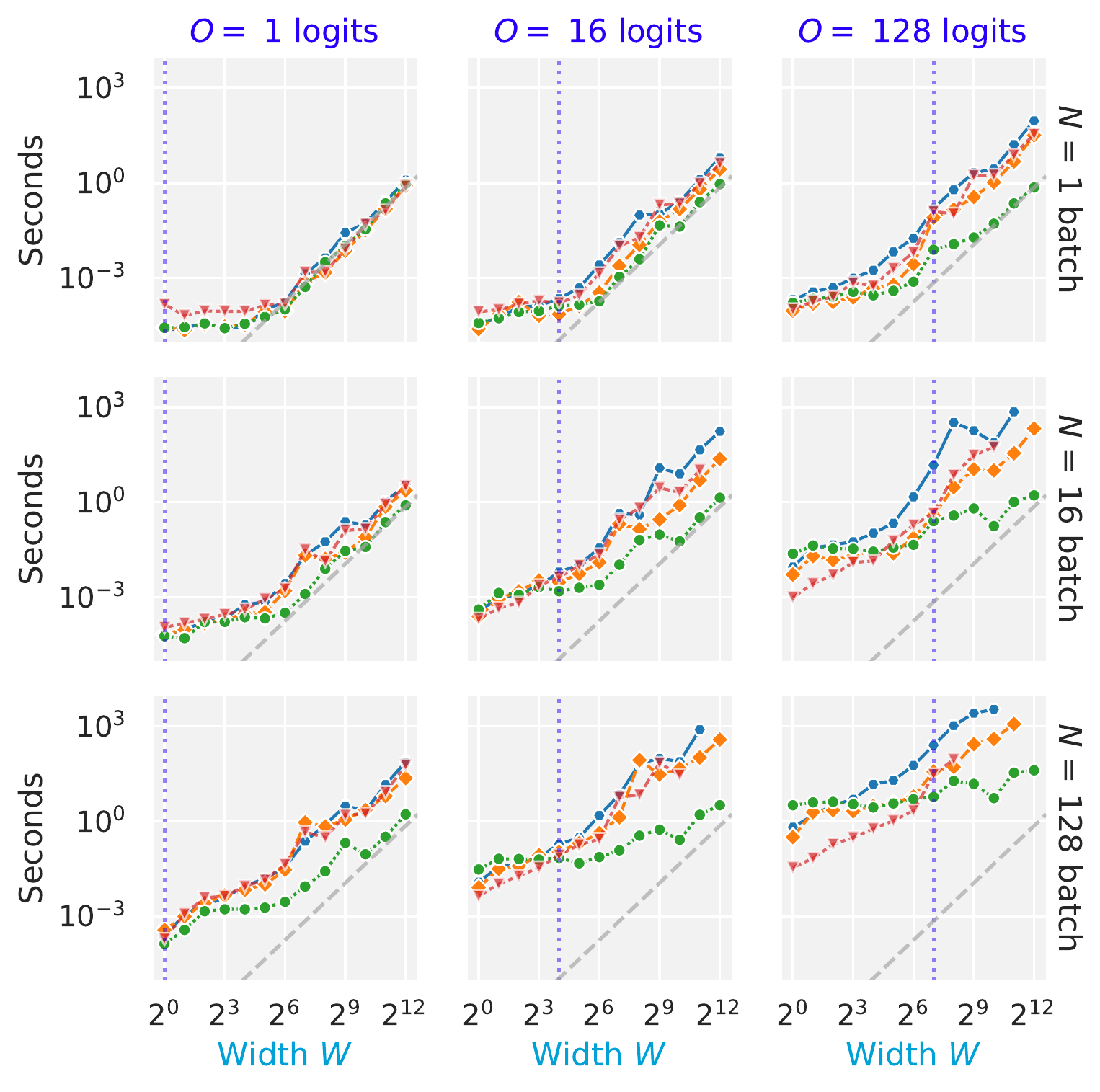}
            \includegraphics[width=0.48\textwidth]{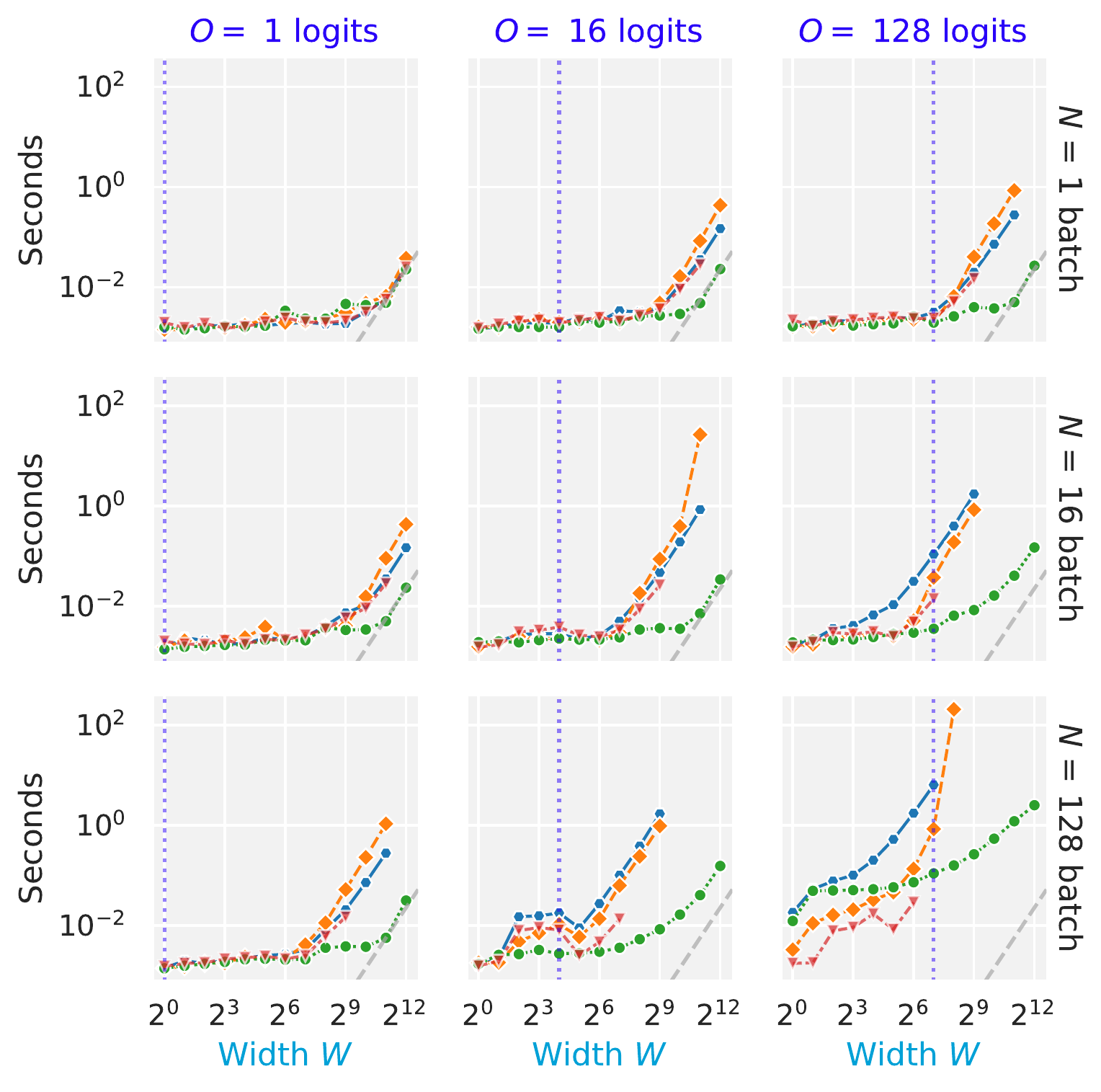}\\
            
            \hfil\hfil\textbf{TPUv4}\hfil\hfil\hfil\hfil\hfil\textbf{NVIDIA P100}\hfil\hfil\hfil\\
            \includegraphics[width=0.48\textwidth]{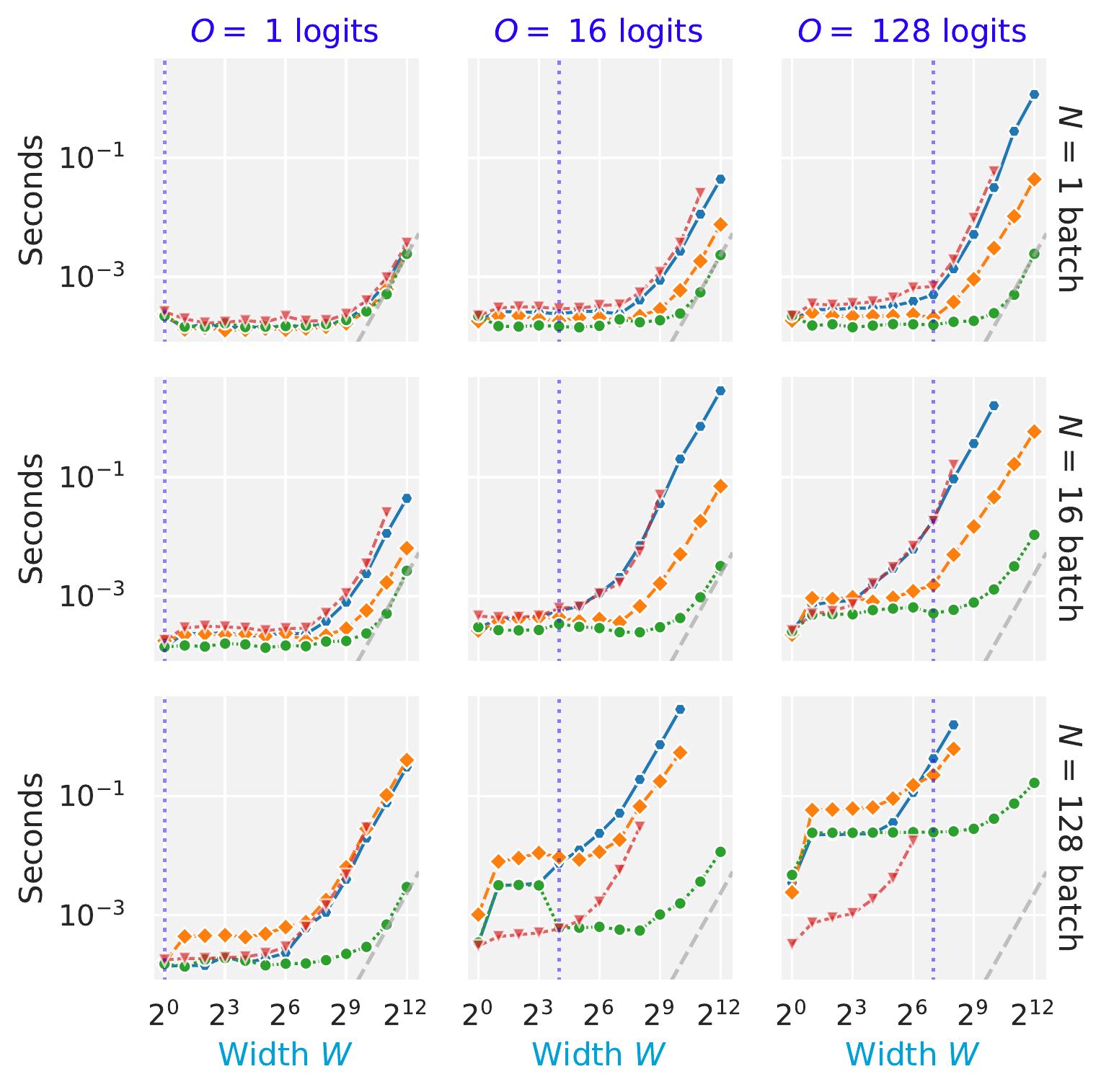}
            \includegraphics[width=0.48\textwidth]{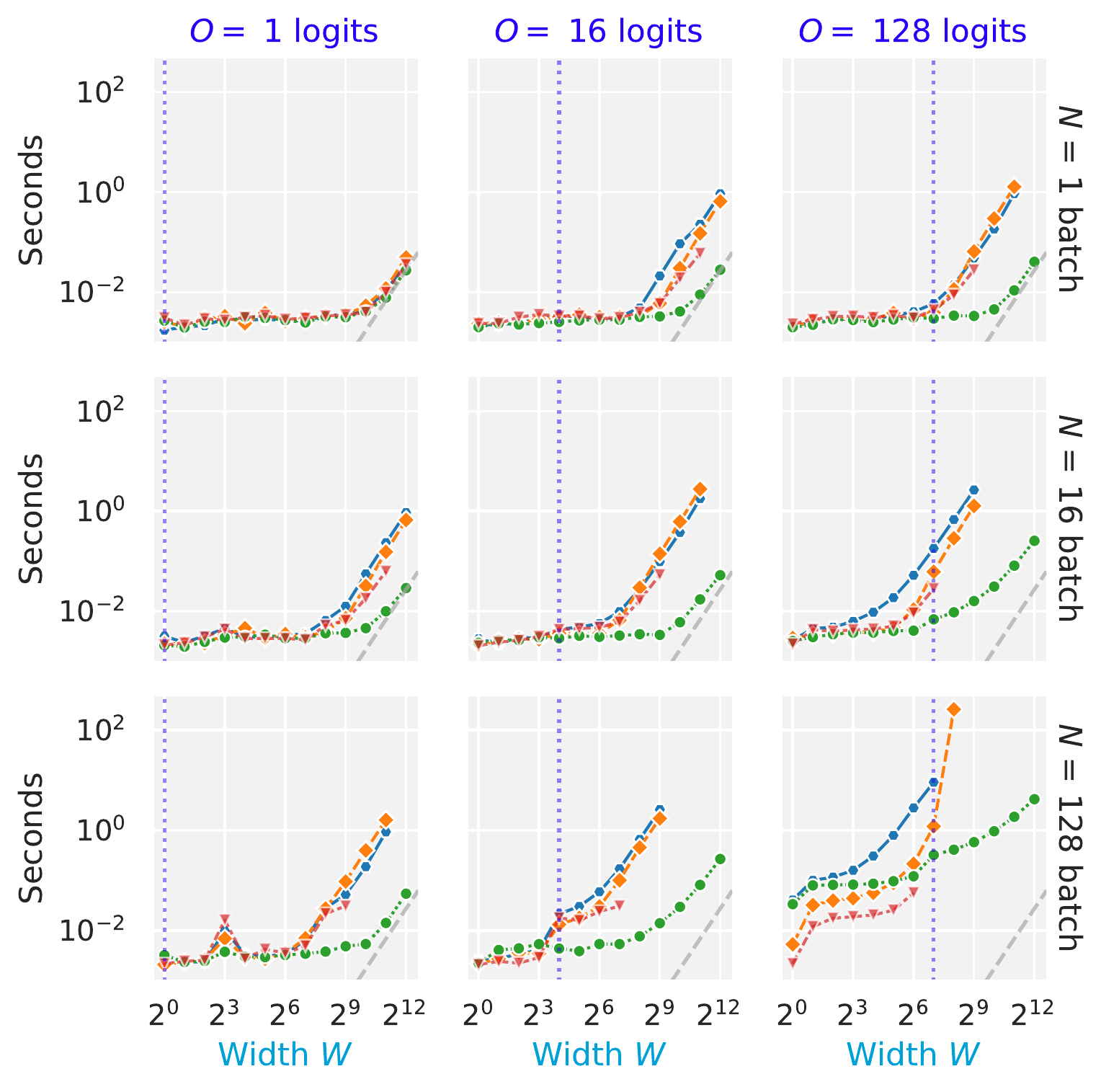}
            
            \vspace{0.2cm}
            
            \hphantom{12345\,\,}\includegraphics[width=0.9\textwidth]{figures/legend.pdf}
            
            \caption{
                \textbf{Wall-clock time of computing the NTK of a 10-layer ReLU FCN on different platforms.} In all settings, \sd~allow orders of magnitude improvement in wall-clock time and memory (missing points indicate out-of-memory error). However, we remark that on GPU platforms (right), \ntvp~deliver a robust improvement only for large $\o$ (rightmost column), while for $\o = 16$ the cost is comparable or even larger than \jc. See \cref{fig:fcn_flops} for FLOPs, \textbf{TPUv3} platform, and more discussion. See \sref{sec:experimental_definition} for details.}\label{fig:fcn_secs_other}
        \end{figure}
        
        \begin{figure}[H]
            \centering
            \includegraphics[width=\textwidth]{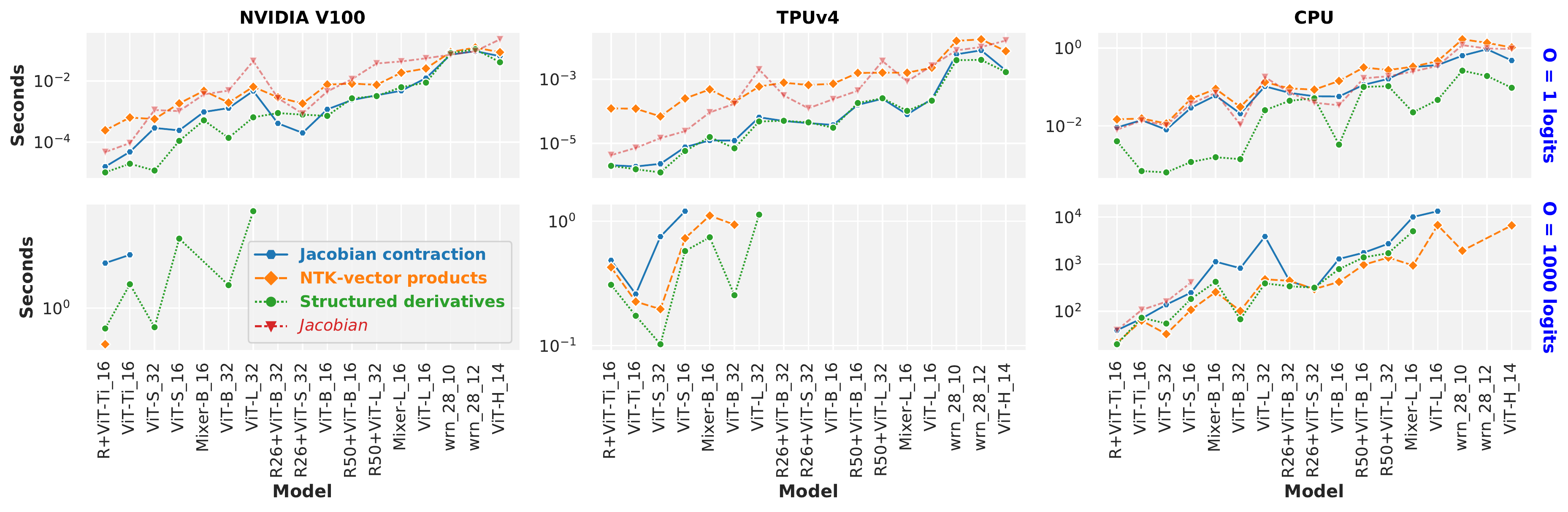}
            \caption[]{
                \textbf{Wall-clock time per input pair of computing NTK on various ImageNet models} like Vision Tansformers and hybrids \citep{dosovitskiy2021an, steiner2021augreg}, WideResNets \citep{zagoruyko2016wide} and MLP-Mixers \citep{tolstikhin2021mlpmixer}. 
                \vspace{0.1cm}
                
                \sd~generally allow fastest computation, but also are able to process more models due to lower memory requirements (lower left; missing points indicate out-of-memory error). For the case of single output logit $\o = 1$ (top row), 
                \ntvp~are generally detrimental due to a costly forward pass \fp~relative to the size of parameters $\p$ (i.e. a lot of weight sharing; see \cref{tab:intro_summary_informal}). However, since \ntvp~scale better than other methods with output size, for $\o = 1000$ (bottom row), they perform comparably or better than other methods. 
                \vspace{0.1cm}
                
                Finally, we remark that the \jac~not only runs out of memory faster, but can also take more time to compute. We conjecture that due to a larger memory footprint, \xla~can sometimes perform optimizations that trade off speed for memory, and therefore compute the \jac~in a less optimal way than if it had more memory available. Alternatively, \xla~could also be performing simplifications of the NTK expression in these cases, such that those would not be possible in \jac~computation alone. 
                \vspace{0.1cm}
                
                See \cref{fig:resnet_secs} for ResNets, and \sref{sec:experimental_definition} for details.
            }\label{fig:imagenet_other_secs}
            
            \vspace{1cm}
        
            \centering
            \includegraphics[width=\textwidth]{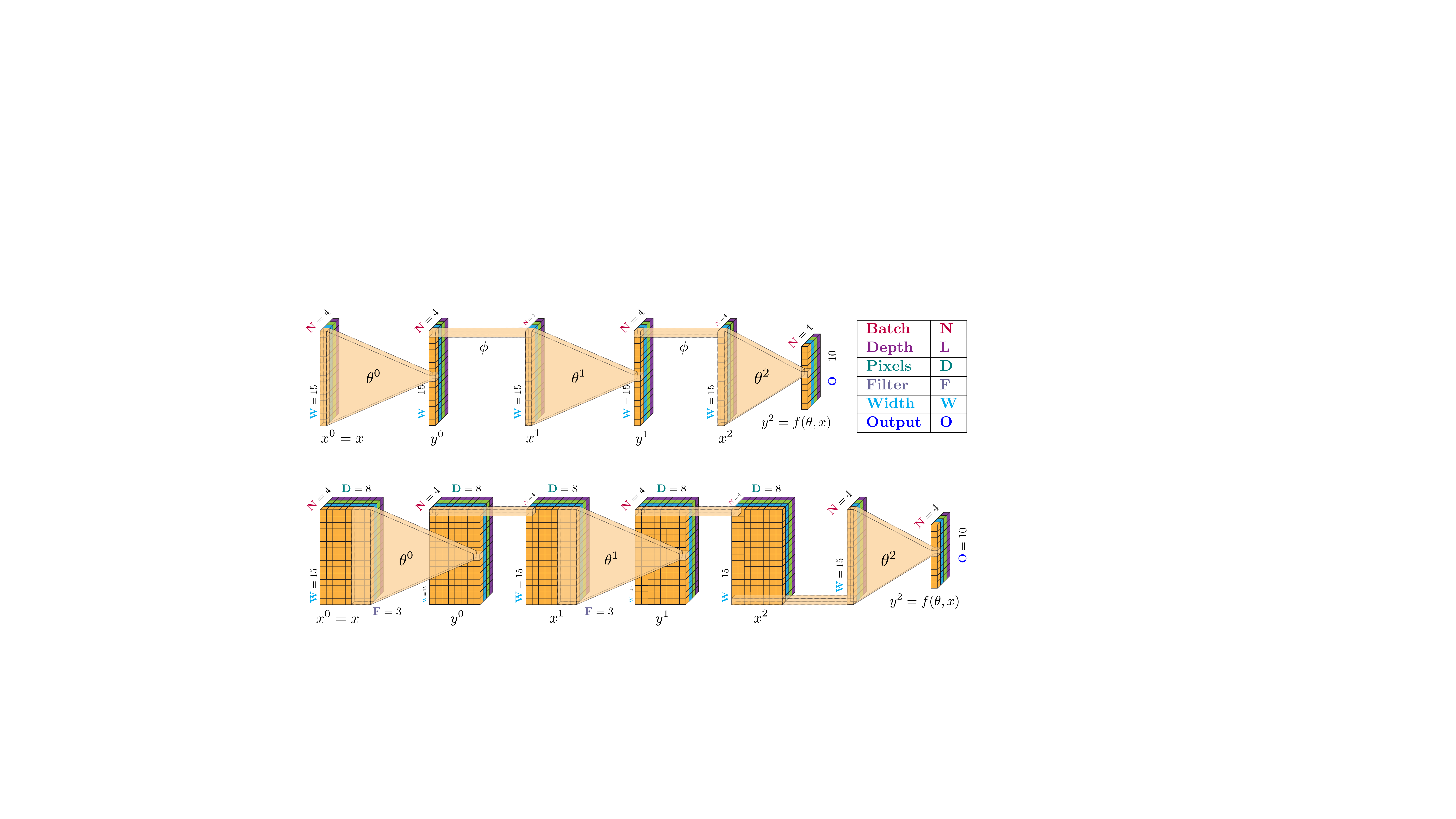}
            \caption{
                \textbf{Notation used in \sref{sec:example_fcn} (FCN, top) and \sref{sec:example_cnn} (CNN, bottom).} In both settings $\t = 2$. For FCN, $\l = 5$ (3 matrix multiplication primitives and two nonlinearities $\phi$), $\d = \f = 1$. For CNN, there is an extra global average pooling primitive as the penultimate layer, therefore $\l = 6$, and $\d = 8$, $\f = 3$.
            }\label{fig:fcn_legend}
        \end{figure}

    \subsection{Glossary}\label{sec:glossary}
    
        \begin{itemize}
            
            \item $\n$ - batch size of inputs $x$ to the NN $f\left(\theta, x\right)$.
                \subitem $\ni$ - batch index ranging from $1$ to $\n$.
            \item $\o$ - output size (e.g. number of logits) of the NN $f\left(\theta, x\right)$ for a single ($\n = 1$) input $x$.
                \subitem $\oi$ - output index ranging from $1$ to $\o$.
            \item The NTK matrix has shape $\n\o\times\n\o$.
            \item $\w$ - width of an FCN, or number of channels of a CNN. Individual inputs $x$ are usually assumed to have the same size / number of channels.
            \item $\t$ - number of trainable parameter matrices, that are used in a possibly different number of primitives in the network. Without weight sharing, synonymous with the depth of the network.
                \subitem $\ti$ - depth index ranging from $0$ to $\t$.
            \item $\l$ - number of primitives (nodes in the computation graph) of the network $f(\theta, x)$. Without weight sharing, synonymous with the depth of the network and is proportional to $\t$.
                \subitem $\li$ - primitive index ranging from $1$ to $\l$.
            \item $\d$ - total number of pixels (e.g. $1024$ for a $32 \times 32$ image; $1$ for an FCN) in an input and every intermediate layer of a CNN (\code{SAME} or \code{CIRCULAR} padding, unit stride, and no dilation are assumed to ensure the spatial size unchanged from layer to layer).
            \item $\f$ - total filter size (e.g. 9 for a $3 \times 3$ filter; $1$ for an FCN) in a convolutional filter of a CNN.
            \item $\y$ - total output size of a primitive $y$ (e.g. $\y = \d\w$ for a layer with $\d$ pixels and $\w$ channels; $\y = \w$ for FCN). Depending on the context, can represent size of a single or particular primitive in the network, or the size of all primitives together.
                \subitem $y$ - intermediate primitive output or intermediate primitive as a function of parameters $y\left(\theta^\ti\right)$, depending on the context.
            \item $\c$ - in \sref{sec:more_structure}, the size of the axis along which a primitive Jacobian $\partial y\big/\partial \theta$ admits certain structure ($\c$ can often be equal to $\y$ or a significant fraction of it, e.g. $\w$).
                \subitem $\ci$ - index along the structured axis, ranging from $1$ to $\c$.
            \item $\p$ - total size of trainable parameters. Depending on the context, can represent the size of a particular weight tensor $\theta^\ti$ in some layer $\ti$ (e.g. $\w^2$ for width-$\w$ FCN), or the size of all parameters in the network.
            \item \fp~- forward pass, cost (time or memory, depending on the context) of evaluating $f(\theta, x)$ on a single ($\n = 1$) input $x$.
            \item \textbf{MJJMP} - matrix-Jacobian-Jacobian-matrix product, an AD operation necessary to evaluate the NTK as in \cref{eq:expanded_ntk_main}, and the respective time and memory cost of the operation. As we describe in \sref{sec:str_derivatives}, our efficient implementation of MJJMPs often allows to evaluate the NTK much faster than when using standard AD operations like JVPs and VJPs. Note that unlike other variables, \textbf{MJJMP} represents the batched (accounting for $\n$) contraction cost, since batched \textbf{MJJMP} can have non-trivial dependence on $\n$.
        
        \end{itemize}

    \subsection{List of Primitives and their Structures}\label{sec:primitives}
    
        \begin{table}[H]
            \centering
            \resizebox{0.92\textwidth}{!}{
                \begin{tabular}{|l|l|l|l|}
                    \hline
                    \textbf{Transposable primitive in \texttt{jax.ad.primitive\_transposes}} & \textbf{\cbd} & \textbf{\bd} & \textbf{\obt} \\ \hline
                    add                                                                   & \checkmark    &              & \checkmark    \\ \hline
                    add\_any                                                              & \checkmark    &              & \checkmark    \\ \hline
                    all\_gather                                                           &               &              &               \\ \hline
                    all\_to\_all                                                          &               &              &               \\ \hline
                    broadcast\_in\_dim                                                    & \checkmark    &              & \checkmark    \\ \hline
                    call                                                                  &               &              &               \\ \hline
                    closed\_call                                                          &               &              &               \\ \hline
                    complex                                                               &               &              &               \\ \hline
                    concatenate                                                           & \checkmark    &              &               \\ \hline
                    conj                                                                  &               &              &               \\ \hline
                    conv\_general\_dilated                                                & \checkmark    & \checkmark   &               \\ \hline
                    convert\_element\_type                                                & \checkmark    &              &               \\ \hline
                    copy                                                                  & \checkmark    &              &               \\ \hline
                    cumsum                                                                &               &              &               \\ \hline
                    custom\_lin                                                           &               &              &               \\ \hline
                    custom\_linear\_solve                                                 &               &              &               \\ \hline
                    custom\_transpose\_call                                               &               &              &               \\ \hline
                    device\_put                                                           & \checkmark    &              &               \\ \hline
                    div                                                                   & \checkmark    & \checkmark   &               \\ \hline
                    dot\_general                                                          & \checkmark    & \checkmark   &               \\ \hline
                    dynamic\_slice                                                        &               &              &               \\ \hline
                    dynamic\_update\_slice                                                &               &              &               \\ \hline
                    fft                                                                   &               &              &               \\ \hline
                    gather                                                                &               &              &               \\ \hline
                    imag                                                                  &               &              &               \\ \hline
                    linear\_call                                                          &               &              &               \\ \hline
                    mul                                                                   & \checkmark    & \checkmark   &               \\ \hline
                    named\_call                                                           &               &              &               \\ \hline
                    neg                                                                   & \checkmark    &              &               \\ \hline
                    pad                                                                   & \checkmark    &              &               \\ \hline
                    pdot                                                                  &               &              &               \\ \hline
                    ppermute                                                              &               &              &               \\ \hline
                    psum                                                                  &               &              &               \\ \hline
                    real                                                                  &               &              &               \\ \hline
                    reduce\_sum                                                           & \checkmark    &              &               \\ \hline
                    reduce\_window\_sum                                                   & \checkmark    &              &               \\ \hline
                    remat\_call                                                           &               &              &               \\ \hline
                    reshape                                                               & \checkmark    &              &               \\ \hline
                    rev                                                                   & \checkmark    &              &               \\ \hline
                    scatter                                                               &               &              &               \\ \hline
                    scatter-add                                                           &               &              &               \\ \hline
                    scatter-mul                                                           &               &              &               \\ \hline
                    select\_n                                                             &               &              &               \\ \hline
                    select\_and\_gather\_add                                              &               &              &               \\ \hline
                    select\_and\_scatter\_add                                             &               &              &               \\ \hline
                    sharding\_constraint                                                  &               &              &               \\ \hline
                    slice                                                                 &               &              &               \\ \hline
                    squeeze                                                               & \checkmark    &              &               \\ \hline
                    sub                                                                   & \checkmark    &              & \checkmark    \\ \hline
                    transpose                                                             & \checkmark    &              &               \\ \hline
                    triangular\_solve                                                     &               &              &               \\ \hline
                    while                                                                 &               &              &               \\ \hline
                    xla\_call                                                             &               &              &               \\ \hline
                    xla\_pmap                                                             &               &              &               \\ \hline
                    xmap                                                                  &               &              &               \\ \hline
                    zeros\_like                                                           & \checkmark    &              &               \\ \hline
                \end{tabular}
            }
            \caption{
                \textbf{List of all linear primitives and currently implemented \sd~rules from \sref{sec:more_structure}.} In the future, more primitives and more rules can be supported, yet at the time of writing even the small set currently covered enables dramatic speed-up and memory savings in contemporary ImageNet models as in \cref{fig:resnet_secs} and \cref{fig:imagenet_other_secs}.
            }\label{tab:list_of_primitives}
        \end{table}

    \subsection{JVP and VJP Costs}\label{sec:jvp_vjp_costs}
        
        Here we provide intuition for why JVP and VJP are asymptotically equivalent in time to the forward pass \fp, as we mentioned in \sref{sec:jvps_vjps_main}. See \citep[Section 3]{evaluating_derivatives} and \citep{radul2022you} for a rigorous treatment, and the \href{https://jax.readthedocs.io/en/latest/notebooks/autodiff_cookbook.html}{JAX Autodiff Cookbook} for a hands-on introduction.
    
\begin{figure}
            \centering
            \vspace{-0.15cm}
            \includegraphics[width=\textwidth]{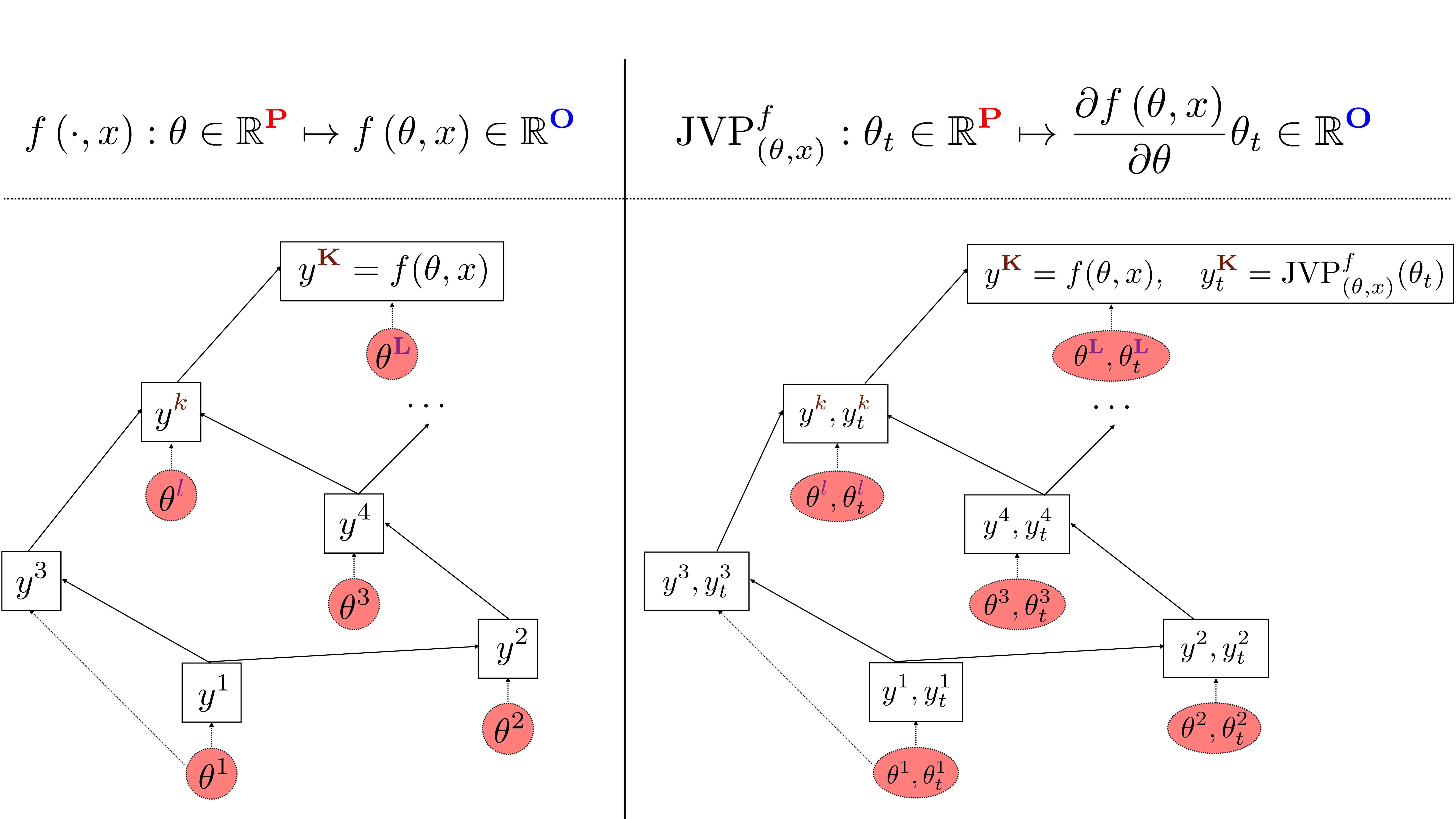}
            \caption{
                \textbf{Visual demonstration for why JVP time and memory costs are asymptotically comparable to the forward pass (\fp).} \textbf{Left:} computational graph of the forward pass $f(\theta, x)$. \textbf{Right:} computational graph of joint evaluation of the forward pass $f(\theta, x)$ along with $\textrm{JVP}^f_{\left(\theta, x\right)}\left(\theta_t\right)$. Each node of the JVP graph accepts both primal and tangent inputs, and returns primal and tangent outputs, but the topology of the graph and order of execution remains identical to \fp. As long as individual nodes of the JVP graph do not differ significantly in time and memory from the \fp~nodes, time and memory of a JVP ends up asymptotically equivalent to \fp~due to identical graph structure. However, in order to create JVP nodes and evaluate them, the time cost does grow by a factor of about 3 compared to \fp. See \sref{sec:jvp_vjp_costs} for discussion.
            }\label{fig:jvp_vjp}
        \end{figure}

        \textbf{JVP} can be computed by traversing a computational graph of the same topology as \fp, except for primitive nodes in the graph need to be augmented to compute not only the forward pass of the node, but also the JVP of the node (see \cref{fig:jvp_vjp}). Due to identical topology and order of evaluation, asymptotically time and memory costs remain unchanged. However, constructing the augmented nodes in the JVP graph, and their consequent evaluation results in extra time cost proportional to the size of the graph. Therefore in practice JVP costs about $3 \times\fp$ time and $2 \times\fp $ memory.
        
        \textbf{VJP}, as a linear function of cotangents $f_c$, is precisely the transpose of the linear function JVP. As such, it can be computed by traversing the transpose of the JVP graph (\cref{fig:jvp_vjp}, right), with each JVP node replaced by its transposition as well. This results in identical time and memory costs, as long as node transpositions are implemented efficiently. However, their evaluation requires primal outputs $y^\li$ (now inputs to the transpose nodes), which is why VJP necessitates an extra \fp~time cost to compute them (hence costlier than JVP, but still inconsequential asymptotically) and extra memory to store them, which can generally increase asymptotic memory requirements.
        
    \subsection{Types of Structured Derivatives and Associated MJJMP Costs}\label{sec:more_structure}
        
        Here we continue \sref{sec:str_derivatives} and list the types of structures in primitive Jacobians $\partial y\big/\partial \theta$ that allow linear algebra simplifications of the respective MJJMPs in \cref{eq:expanded_ntk_main}. Resulting contraction (\textbf{MJJMP}) and memory ($\j$) costs from the following subsections are summarized in \cref{tab:complexities_structures}, and the list of primitives annotated with their types of structure is presented in \cref{tab:list_of_primitives}.

        \begin{table}
            \centering
            \resizebox{\textwidth}{!}{
                \begin{tabular}{|l|l|l|l|l|}
                    \hline
                    \multirow{2}{*}{Structure of $\partial y\big/\partial \theta \downarrow$}                                                     & \multicolumn{3}{c|}{\textbf{MJJMP} (time; minimum of the 3 values)}                                                                                      & \multirow{2}{*}{$\n\j$ (memory)}                                      
                    \\ 
                    \cline{2-4}                                             & \textbf{Outside-in}                                                 & \textbf{Left-to-right}                                                   & \textbf{Inside-out}                                                           &
                    \\ \hline
                    \hyperref[sec:str_derivatives]{None w/ VJPs \& JVPs}    & $\n \o \left[\fp\right]\hphantom{0\,\,} + \n^2 \o^2 \p$             & $\n^2 \o \left[\fp\right]$                                               & Not possible                                                                  & 0                
                    \\ \hline
                    \hyperref[sec:default_cost]{None w/ explicit matrices}  & $\n \o \y \p\hphantom{/\c^2} + \n^2 \o^2 \p$                        & $\n^2 \o \y \p\hphantom{/\c^2} + \n^2\o^2 \y$                            & $\n^2 \y^2 \p\hphantom{/\c^2} + \n^2\o \y^2\hphantom{/\c^2} + \n^2\o^2 \y$    & $\n\y\p$             
                    \\ \hline
                    \bd                                                     & $\n \o \y \p/\c\hphantom{^2} + \n^2 \o^2 \p$                        & $\n^2 \o \y \p/\c\hphantom{^2} + \n^2\o^2 \y$                            & $\n^2 \y^2 \p/\c^2 + \n^2\o \y^2/\c\hphantom{^2} + \n^2\o^2 \y$               & $\n\y\p/\c^2$          
                    \\ \hline
                    \cbd                                                    & $\n \o \y \p/\c\hphantom{^2} + \n^2 \o^2 \p$                        & $\n^2 \o \y \p/\c\hphantom{^2} + \n^2\o^2 \y$                            & $\n^2 \y^2 \p/\c^3 + \n^2\o \y^2/\c\hphantom{^2} + \n^2\o^2 \y$               & $\n\y\p/\c^2$            
                    \\ \hline
                    \ibt                                                    & $\n \o \y \p/\c\hphantom{^2} + \n^2 \o^2 \p$                        & $\n^2 \o \y \p/\c\hphantom{^2} + \n^2\o^2 \y$                            & $\n^2 \y^2 \p/\c\hphantom{^2} + \n^2\o \y^2\hphantom{/\c^2} + \n^2\o^2 \y$    & $\n\y\p/\c$ 
                    \\ \hline
                    \obt                                                    & $\n \o \y \p/\c\hphantom{^2} + \n^2 \o^2 \p\hphantom{/\c} + \n\o\y$ & $\n^2 \o \y \p/\c\hphantom{^2} + \n^2\o^2 \y / \c\hphantom{^2} + \n\o\y$ & $\n^2 \y^2 \p/\c^2 + \n^2\o \y^2/\c^2 + \n^2\o^2 \y / \c + \n\o \y$           & $\n\y\p/\c$ 
                    \\ \hline
                    \bt                                                     & $\n \o \y \p/\c^2 + \n^2 \o^2 \p/\c + \n\o\y$                       & $\n^2 \o \y \p/\c^2 + \n^2\o^2 \y / \c^2 + \n\o\y$                       & $\n^2 \y^2 \p/\c^3 + \n^2\o \y^2/\c^2 + \n^2\o^2 \y / \c + \n\o \y$           &  $\n\y\p/\c^2$
                    \\ \hline
                \end{tabular}
            }
            \caption[]{
                \textbf{Asymptotic time (MJJMP) and extra memory ($\n\j$) costs of computing the contractions for NTK summands $\nt$ of shape $\n \o\times \n\o$ in \sref{sec:str_derivatives} and \cref{tab:intro_summary_informal}.} Time complexity of the MJJMP in \sd~is the minimum (due to using \code{np.einsum} with optimal contraction order) of the 3 time entries in the row corresponding to the structure present in a pair of primitives $y_1^{\li_1}$ and $y_2^{\li_2}$. How it compares to \jc~and \ntvp~(top row) depends on the specific primitive, notably the cost of evaluating the primitive \fp. See \cref{tab:intro_summary} and \cref{tab:intro_summary_1x1} for exact comparison in the case of matrix multiplication and convolution. See \sref{sec:glossary} for legend.
            }\label{tab:complexities_structures}
        \end{table}

        \subsubsection{No Structure}\label{sec:default_cost}
        
            We first consider the default cost of evaluating a single summand in \cref{eq:expanded_ntk_main}, denoting individual matrix shapes underneath:
            \begin{equation}\label{eq:contraction}
                \nt 
                = \frac{\partial f_1}{\partial y_1^{\li_1}}\frac{\partial y_1^{\li_1}}{\partial \theta^{\ti}}\frac{\partial y_2^{\li_2}}{\partial \theta^{\ti}}^T\frac{\partial f_2}{\partial y_2^{\li_2}}^T 
                \eqqcolon \overbrace{\underbrace{\frac{\partial f_1}{\partial y_1}}_{\o \times \y}
                \quad
                \underbrace{\frac{\partial y_1}{\partial \theta\vphantom{y_1^{i_1}}}}_{\y\times \p}
                \quad
                \underbrace{\frac{\partial y_2}{\partial \theta\vphantom{y_1}}^T}_{\p\times \y}
                \quad
                \underbrace{\frac{\partial f_2}{\partial y_2}^T}_{\y\times \o}}^{\o \times \o}
            \end{equation}
            We have dropped indices $\ti$, $\li_1$ and $\li_2$ on the right-hand side of \cref{eq:contraction} to avoid clutter, and consider $\theta \coloneqq \theta^{\ti}, y_1 \coloneqq y_1^{\li_1}, y_2 \coloneqq y_2^{\li_2}$ until the end of this section. There are 3 ways of contracting \cref{eq:contraction} that cost
            \begin{enumerate}
            	\item[(a)] \textbf{Outside-in:} $\o \y \p + \o^2 \p$ 
            	\item[(b)] \textbf{Left-to-right and right-to-left:} $\o \y \p + \o^2 \y$.
            	\item[(c)] \textbf{Inside-out-left and inside-out-right:} $\y^2 \p + \o \y^2 + \o^2 \y$. 
            \end{enumerate}
            In the next sections, we look at how these costs are reduced given certain structure in $\partial y\big/\partial \theta$.

        \subsubsection{Block Diagonal}\label{sec:block_diagonal}
        	
        	Assume ${\partial y}/{\partial \theta} = \oplus_{\ci=1}^{\c}{\partial y^{\ci}}/{\partial \theta_{\ci}}$, where $\oplus$ stands for \href{https://en.wikipedia.org/wiki/Matrix_addition#Direct_sum}{direct sum of matrices}, i.e. ${\partial y}/{\partial \theta}$ is a block diagonal matrix made of blocks $\left\{{\partial y^{\ci}}/{\partial \theta_{\ci}}\right\}_{\ci=1}^{\c}$, where ${\partial y^{\ci}}/{\partial \theta_{\ci}}$ have shapes $(\y / \c) \times (\p / \c)$. Here $\left\{y^{\ci}\right\}_{\ci=1}^{\c}$ and $\left\{\theta_{\ci}\right\}_{\ci=1}^{\c}$ are \href{https://en.wikipedia.org/wiki/Partition_of_a_set}{partitions} of $y$ and $\theta$ respectively. In NNs this structure is present in binary bilinear operations (on $\theta$ and another argument) such as multiplication, division, batched matrix multiplication, or depthwise convolution. Then \cref{eq:contraction} can be re-written as
        	\begin{align}
        		\nt 
        		&=\frac{\partial f_1}{\partial y_1} \frac{\partial y_1}{\partial \theta}\frac{\partial y_2}{\partial \theta}^T\frac{\partial f_2}{\partial y_2}^T \\ 
        		&=\frac{\partial f_1}{\partial y_1} \left(\oplus_{\ci=1}^{\c}\frac{\partial y^{\ci}_1}{\partial \theta_{\ci}}\right)\left(\oplus_{\ci=1}^{\c}\frac{\partial y^{\ci}_2}{\partial \theta_{\ci}}\right)^T\frac{\partial f_2}{\partial y_2}^T \\
        		&=\frac{\partial f_1}{\partial y_1} \left(\oplus_{\ci=1}^{\c} \left[\frac{\partial y_1^{\ci}}{\partial \theta_{\ci}}\frac{\partial y_2^{\ci}}{\partial\theta_{\ci}}^T\right]\right)\frac{\partial f_2}{\partial y_2}^T \\
        		&=\sum_{\ci=1}^{\c} \frac{\partial f_1}{\partial y_1^{\ci}} \left[\frac{\partial y_1^{\ci}}{\partial \theta_{\ci}}\frac{\partial y_2^{\ci}}{\partial \theta_{\ci}}^T\right]\frac{\partial f_2}{\partial y_2^{\ci}}^T,
        	\end{align}
        	where we have applied the block matrix identity 
        	\begin{equation}\label{eq:direct_sum}
        	    [A^1, \dots, A^\c]^T \left(\oplus_{\ci=1}^{\c} B^\ci\right) [D^1, \dots, D^\c] = \sum_{\ci=1}^{\c}A^\ci B^\ci D^\ci.
        	\end{equation}
        	We now perform a complexity analysis similar to \cref{eq:contraction}:
        	$$
        	    \nt = 
        	    \sum_{\ci=1}^{\c}
        	    \quad
        	    \overbrace{\underbrace{\frac{\partial f_1}{\partial y_1^{\ci}}}_{\o \times (\y/\c)}
        	    \quad
        	    \underbrace{\frac{\partial y_1^{\ci}}{\partial \theta_{\ci}}}_{(\y/\c)\times (\p/\c)}
        	    \quad
        	    \underbrace{\frac{\partial y_2^{\ci}}{\partial \theta_{\ci}}^T}_{(\p/\c)\times (\y/\c)}
        	    \quad
        	    \underbrace{\frac{\partial f_2}{\partial y_2^{\ci}}^T}_{(\y/\c)\times \o}}^{\o \times \o}
        	$$
        	In this case complexities of the three methods become
        	\begin{enumerate}
        		\item \textbf{Outside-in:} $\o \y \p/\c + \o^2 \p$.
        		\item \textbf{Left-to-right and right-to-left:} $\o \y \p/\c + \o^2 \y$.
        		\item \textbf{Inside-out-left and inside-out-right:} $\y^2 \p/\c^2 + \o \y^2/\c + \o^2 \y$.
        	\end{enumerate}

        \subsubsection{Constant-block Diagonal}\label{sec:constant_block_diagonal} 
        	
        	Assume $\frac{\partial y}{\partial \theta} = I_{\c} \otimes \frac{\partial y^1}{\partial \theta_1}$, and $\frac{\partial y^1}{\partial \theta_1}$ has shape $(\y / \c) \times (\p / \c)$. In NNs, this is present in fully-connected, convolutional, locally-connected, attention, and many other layers that contain a matrix multiplication along some axis. This is also present in all unary elementwise linear operations on $\theta$ like transposition, negation, reshaping and many others. This is a special case of \sref{sec:block_diagonal} with $\frac{\partial y^{\ci}}{\partial \theta_{\ci}} = \frac{\partial y^1}{\partial \theta_1}$ for any $\ci$. Here a similar analysis applies, yielding 
        	$$
        	    \nt = 
        	    \sum_{\ci=1}^{\c}
        	    \quad
        	    \overbrace{\underbrace{\frac{\partial f_1}{\partial y_1^{\ci}}}_{\o \times (\y/\c)}
        	    \quad
        	    \underbrace{\frac{\partial y_1^1}{\partial \theta_1}}_{(\y/\c)\times (\p/\c)}
        	    \quad
        	    \underbrace{\frac{\partial y_2^1}{\partial \theta_1}^T}_{(\p/\c)\times (\y/\c)}
        	    \quad
        	    \underbrace{\frac{\partial f_2}{\partial y_2^{\ci}}^T}_{(\y/\c)\times \o}}^{\o \times \o}
        	$$
        	and the same contraction complexities as in \sref{sec:block_diagonal}, except for the \textbf{Inside-out} order, where the inner contraction term costs only $\y^2\p/\c^3$, since it is only contracted once instead of $\c$ times as in the \bd~case.

        \subsubsection{Input Block-tiled}\label{sec:input_block_tiled}
        
            Assume $\frac{\partial y}{\partial \theta} = \mathbbm{1}_{(1, \c)} \otimes \frac{\partial y}{\partial \theta_1}$, where $\mathbbm{1}_{(1, \c)}$ is an \href{https://en.wikipedia.org/wiki/Matrix_of_ones}{all-ones matrix} of shape $1 \times \c$, and $\frac{\partial y}{\partial \theta_1}$ has shape $\y \times (\p /\c)$. This occurs, for example, in summation or taking the mean.
        	\begin{align}
        		\nt 
        		&=\frac{\partial f_1}{\partial y_1} \frac{\partial y_1}{\partial \theta}\frac{\partial y_2}{\partial \theta}^T\frac{\partial f_2}{\partial y_2}^T \\
        		&=\frac{\partial f_1}{\partial y_1} \left(\mathbbm{1}_{(1, \c)} \otimes \frac{\partial y_1}{\partial \theta_1}\right)\left(\mathbbm{1}_{(1, \c)} \otimes \frac{\partial y_2}{\partial \theta_1}\right)^T\frac{\partial f_2}{\partial y_2}^T \\
        		&=\frac{\partial f_1}{\partial y_1} \left(\c \mathbbm{1}_{(1, 1)}\otimes \left[\frac{\partial y_1}{\partial \theta_1}\frac{\partial y_2}{\partial\theta_1}^T\right]\right)\frac{\partial f_2}{\partial y_2}^T \\
        		&=\c\frac{\partial f_1}{\partial y_1} \left[\frac{\partial y_1}{\partial \theta_1}\frac{\partial y_2}{\partial\theta_1}^T\right]\frac{\partial f_2}{\partial y_2}^T.
        	\end{align}
        	The matrix shapes are
        	$$
        	    \nt = \c
        	    \quad
        	    \overbrace{\underbrace{\frac{\partial f_1}{\partial y_1}}_{\o \times \y}
        	    \quad
        	    \underbrace{\frac{\partial y_1}{\partial \theta_1}}_{\y\times (\p/\c)}
        	    \quad
        	    \underbrace{\frac{\partial y_2}{\partial \theta_1}^T}_{(\p/\c)\times \y}
        	    \quad
        	    \underbrace{\frac{\partial f_2}{\partial y_2}^T}_{\y\times \o}}^{\o \times \o}
        	$$
        	Which leads to the following resulting complexities:
        	\begin{enumerate}
        		\item \textbf{Outside-in:} $\o \y \p/\c + \o^2 \p$.
        		\item \textbf{Left-to-right and right-to-left:} $\o \y \p/\c + \o^2 \y$.
        		\item \textbf{Inside-out and inside-out-right:} $\y^2 \p/\c + \o \y^2 + \o^2 \y$.
        	\end{enumerate}

        \subsubsection{Output Block-tiled}\label{sec:output_block_tiled}
        
            Assume $\frac{\partial y}{\partial \theta} = \mathbbm{1}_{(\c, 1)} \otimes \frac{\partial y^1}{\partial \theta}$,  where $\frac{\partial y^1}{\partial \theta}$ has shape $(\y / \c) \times \p$. This occurs during broadcasting or broadcasted arithmetic operations. In this case
        	\begin{align}
        		\nt 
        		&=\frac{\partial f_1}{\partial y_1} \frac{\partial y_1}{\partial \theta}\frac{\partial y_2}{\partial \theta}^T\frac{\partial f_2}{\partial y_2}^T \\
        		&=\frac{\partial f_1}{\partial y_1} \left(\mathbbm{1}_{(\c, 1)} \otimes \frac{\partial y^1_1}{\partial \theta}\right)\left(\mathbbm{1}_{(\c, 1)} \otimes \frac{\partial y^1_2}{\partial \theta}\right)^T\frac{\partial f_2}{\partial y_2}^T \\
        		&=\frac{\partial f_1}{\partial y_1} \left(\mathbbm{1}_{(\c,\c)}\otimes \left[\frac{\partial y_1^1}{\partial \theta}\frac{\partial y_2^1}{\partial\theta}^T\right]\right)\frac{\partial f_2}{\partial y_2}^T \\
        		&=\left(\sum_{\ci=1}^{\c} \frac{\partial f_1}{\partial y_1^{\ci}}\right) \left[\frac{\partial y_1^1}{\partial \theta_1}\frac{\partial y_2^1}{\partial\theta_1}^T\right]\left(\sum_{\ci=1}^{\c}\frac{\partial f_2}{\partial y_2^{\ci}}^T\right),
        	\end{align}
        	where we have used a block matrix identity
        	$$
        	    [A^1, \dots, A^\c]^T \left(\mathbbm{1}_{(\c, \c)} \otimes B\right) [D^1, \dots, D^\c] = \left(\sum_{\ci=1}^{\c}A^\ci\right) B \left(\sum_{\ci=1}^{\c}D^\ci\right).
        	$$
        	
        	Finally, denoting the shapes,
        	$$
        	    \nt = 
        	    \overbrace{\underbrace{\left(\sum_{\ci=1}^{\c} \frac{\partial f_1}{\partial y_1^{\ci}}\right)}_{\o \times (\y/\c)}
        	    \quad
        	    \underbrace{\frac{\partial y_1^1}{\partial \theta}}_{(\y/\c)\times \p}
        	    \quad
        	    \underbrace{\frac{\partial y_2^1}{\partial \theta}^T}_{\p\times (\y/\c)}
        	    \quad
        	    \underbrace{\left(\sum_{\ci=1}^{\c} \frac{\partial f_2}{\partial y_2^{\ci}}^T\right)}_{(\y/\c)\times \o}}^{\o \times \o},
        	$$
        	complexities of the three methods become (notice we add an $\o \y$ term to perform the sums)
        	\begin{enumerate}
        		\item \textbf{Outside-in:} $\o \y \p/\c + \o^2 \p+ \o\y$.
        		\item \textbf{Left-to-right:} $\o \y \p/\c + \o^2 \y / \c + \o\y$.
        		\item \textbf{Inside-out:} $\y^2 \p/\c^2 + \o \y^2/\c^2 + \o^2 \y / \c + \o \y$.
        	\end{enumerate}

        \subsubsection{Block-tiled}\label{sec:block_tiled}
        
            Assume $\frac{\partial y}{\partial \theta} = \mathbbm{1}_{(\c, \c)} \otimes \frac{\partial y^1}{\partial \theta_1}$, where $\frac{\partial y^1}{\partial \theta_1}$ has shape $(\y / \c) \times (\p/\c)$. This occurs for instance when $y$ is a constant. In this case
        	\begin{align}
        		\nt 
        		&= \frac{\partial f_1}{\partial y_1} \frac{\partial y_1}{\partial \theta}\frac{\partial y_2}{\partial \theta}^T\frac{\partial f_2}{\partial y_2}^T \\
        		&=\frac{\partial f_1}{\partial y_1} \left(\mathbbm{1}_{(\c, \c)} \otimes \frac{\partial y^1_1}{\partial \theta_1}\right)\left(\mathbbm{1}_{(\c, \c)} \otimes \frac{\partial y^1_2}{\partial \theta_1}\right)^T\frac{\partial f_2}{\partial y_2}^T \\
        		&=\frac{\partial f_1}{\partial y_1} \left(\c \mathbbm{1}_{(\c,\c)}\otimes \left[\frac{\partial y_1^1}{\partial \theta_1}\frac{\partial y_2^1}{\partial\theta_1}^T\right]\right)\frac{\partial f_2}{\partial y_2}^T \\
        		&=\c\left(\sum_{\ci=1}^{\c} \frac{\partial f_1}{\partial y_1^{\ci}}\right) \left[\frac{\partial y_1^1}{\partial \theta_1}\frac{\partial y_2^1}{\partial\theta_1}^T\right]\left(\sum_{\ci=1}^{\c}\frac{\partial f_2}{\partial y_2^{\ci}}^T\right),
        	\end{align}
        	This results in the following contraction:
        	$$
        	    \nt = \c
        	    \quad
        	    \overbrace{\underbrace{\left(\sum_{\ci=1}^{\c} \frac{\partial f_1}{\partial y_1^{\ci}}\right)}_{\o \times (\y/\c)}
        	    \quad
        	    \underbrace{\frac{\partial y_1^1}{\partial \theta_1}}_{(\y/\c)\times (\p/\c)}
        	    \quad
        	    \underbrace{\frac{\partial y_2^1}{\partial \theta_1}^T}_{(\p/\c)\times (\y/\c)}
        	    \quad
        	    \underbrace{\left(\sum_{\ci=1}^{\c} \frac{\partial f_2}{\partial y_2^{\ci}}^T\right)}_{(\y/\c)\times \o}}^{\o \times \o},
        	$$
        	with final complexities of
        	\begin{enumerate}
        		\item \textbf{Outside-in:} $\o \y \p/\c^2 + \o^2 \p+ \o\y$.
        		\item \textbf{Left-to-right:} $\o \y \p/\c^2 + \o^2 \y / \c^2 + \o\y$.
        		\item \textbf{Inside-out:} $\y^2 \p/\c^3 + \o \y^2/\c^2 + \o^2 \y / \c + \o \y$.
        	\end{enumerate}

        \subsubsection{Batched NTK Cost Analysis}\label{sec:batched_cost_summary}
        
            For simplicity, above we have considered evaluating the NTK for $\n = 1$. In the more general case of inputs of batch sizes $\n_1$ and $\n_2$, the same argument as in previous section follows, but the cost of contractions involving terms from different batches grow by a multiplicative factor of $\n_1\n_2$, while all other costs grow by a factor of $\n_1$ or $\n_2$. To declutter notation we consider $\n_1 = \n_2 = \n$, and summarize resulting costs in \cref{tab:complexities_structures}.

        \subsubsection{Complex Structure Cost Analysis}\label{sec:complex_structure}
            
            In previous sections we have considered $\partial y_1\big/\partial \theta$ and $\partial y_2\big/\partial \theta$ admitting the same, and at most one kind of structure. While this is a common case, in general these Jacobians may admit multiple types of structures along multiple axes (for instance, addition is \cbd~along non-broadcasted axes, and \obt~along the broadcasted axes), and $\partial y_1\big/\partial \theta$ and $\partial y_2\big/\partial \theta$ may have different types of structures and respective axes, if the same weight $\theta$ is used in multiple different primitives of different kind. In such cases, equivalent optimizations are possible (and are implemented in the code) along the largest common subsets of axes for each type of structure that $\partial y_1\big/\partial \theta$ and $\partial y_2\big/\partial \theta$ have. 
            
            For example, let $\theta$ be a matrix in $\mathbb{R}^{\w\times\w}$, $y_1$ be multiplication by a scalar $y_1(\theta) = 2 \theta$, and $y_2$ be matrix-vector multiplication $y_2(\theta) = \theta x$, $x\in \mathbb{R}^{\w}$. In this case $\partial y_1\big/\partial\theta = 2 I_{\w} \otimes I_{\w}$, i.e. it is \cbd~along axes 1 and 2. $\partial y_2\big/\partial\theta = I_{\w}\otimes x^T$, i.e. it is also \cbd, but only along axis 1. Hence, the NTK term containing $\partial y_1\big/\partial \theta$ and $\partial y_2\big/\partial \theta$ will be computed with \cbd~simplification along axis 1. There are probably more computationally optimal ways of processing different structure combinations, as well as more types of structures to be leveraged for NTK computation, and we intend to investigate it in future work.

        \subsubsection{Example: FCN Layer}\label{sec:example_fcn_mjjmp}
        
            In \sref{sec:example_fcn} we have derived the $\j$ and \textbf{MJJMP} costs for an FCN in an improvised manner. Here we arrive at identical complexities using the framework in \sref{sec:more_structure} and \cref{tab:complexities_structures}. Precisely, per \cref{tab:list_of_primitives}, matrix-vector multiplication (\code{dot\_general}) has \cbd~structure with $\c = \y = \w$. Substituting it (along with $\p = \w^2$, $\fp = \w^2$), we obtain \cref{tab:complexities_fcn}, which reproduces our conclusions in \sref{sec:example_fcn} and \cref{tab:intro_summary} for a single layer $\ti < \t$.

            \begin{table*}
                \centering
                \resizebox{\textwidth}{!}{
                    \begin{tabular}{|l|l|l|l|l|}
                        \hline
                        \multirow{2}{*}{Structure of $\partial y\big/\partial \theta \downarrow$}              & \multicolumn{3}{c|}{\textbf{MJJMP} (time; minimum of the 3 values)}                                & \multirow{2}{*}{$\n\j$ (memory)}                                      
                        \\ 
                        \cline{2-4}                                                                            & \textbf{Outside-in}           & \textbf{Left-to-right}           & \textbf{Inside-out}             &                                     
                        \\ \hline
                        \hyperref[sec:default_cost]{None w/ JVPs and VJPs}                                     & $\n^2\o^2\w^2$                & \cellcolor{green!10}$\n^2\o\w^2$ & Not possible                    & 0         
                        \\ \hline
                        \cbd                                                                                   & $\n^2\o^2\w^2$                & $\n^2\o\w^2  + \n^2\o^2\w$       &\cellcolor{green!10}$\n^2\o^2\w$ & $\n\w$
                        \\ \hline
                    \end{tabular}
                }\vspace{0.1cm}
                \caption[]{
                    \textbf{Asymptotic time complexities of computing a single FCN layer NTK contribution}. See \sref{sec:example_fcn_mjjmp} for discussion, \cref{tab:complexities_structures} for a more general setting, \cref{tab:intro_summary} for the case of deep networks, and \sref{sec:glossary} for detailed legend.
                }\label{tab:complexities_fcn}
            \end{table*}

        \subsubsection{Example: CNN Layer}\label{sec:example_cnn_mjjmp}
            
            Per \cref{tab:list_of_primitives} convolution (\code{conv\_general\_dilated}) also has \cbd~structure along the channel axis with $\c = \w$. Substituting it (along with $\p = \f\w^2$, $\fp = \d\f\w^2$, $\y=\d\w$), we obtain \cref{tab:complexities_cnn}, which we next use in \sref{sec:example_cnn}.

            \begin{table*}
                \centering
                \resizebox{\textwidth}{!}{
                    \begin{tabular}{|l|l|l|l|l|}
                        \hline
                        \multirow{2}{*}{Structure of $\partial y\big/\partial \theta \downarrow$}              & \multicolumn{3}{c|}{\textbf{MJJMP} (time; minimum of the 3 values)}                               & \multirow{2}{*}{$\n\j$ (memory)}                                      
                        \\ 
                        \cline{2-4}                                                   & \textbf{Outside-in}                 & \textbf{Left-to-right}           & \textbf{Inside-out}                               &                                     
                        \\ \hline
                        \hyperref[sec:default_cost]{None w/ JVPs and VJPs}            & $\n^2\o^2\f\w^2 + \n\o\d\f\w^2$     & $\n^2\o\d\f\w^2$                 & Not possible                                      & 0          
                        \\ \hline
                        \cbd                                                          & $\n^2\o^2\f\w^2 + \n\o\d\f\w^2$     & $\n^2\o\d\f\w^2 + \n^2\o^2\d\w$  & $\n^2\d^2\f\w + \n^2\o \d^2\w + \n^2\o^2 \d\w$    & $\n\d\f\w$
                        \\ \hline
                    \end{tabular}
                    }
                    \caption[]{
                        \textbf{Asymptotic time complexities of computing a single CNN layer NTK contribution}. See \sref{sec:example_cnn_mjjmp} for discussion, \cref{tab:complexities_structures} for a more general setting, \cref{tab:intro_summary_1x1} for the case of deep networks, and \sref{sec:glossary} for detailed legend.
                }\label{tab:complexities_cnn}\vspace{-0.3cm}
            \end{table*}

    \subsection{CNN NTK Complexity}\label{sec:example_cnn}
    
        Here we derive the CNN NTK complexity for \sref{sec:example_cnn_main_text}, identically to \sref{sec:example_fcn}. We assume the same setup, except for denoting the total filter size as $\f$, total number of pixels $\d$ (unchanged from layer to layer, assuming \code{SAME} or \code{CIRCULAR} padding and unit stride, no dilation), and adding an extra global average pooling layer before the final readout layer (\cref{fig:fcn_legend}, bottom). Note that FCN (\sref{sec:example_fcn}) is a particular case of CNN with $\d=\f=1$.
    
        In this setting $\l = 2\t + 2$ (one more than in \sref{sec:example_fcn}, due to the extra global average pooling primitive). $\p^\ti = \f\w^2$ for $\ti < \t$ and $\o\w$ for $\ti = \t$, $\p = \t\f\w^2 + \o\w$, $\y^\li = \d\w$ for $\li < \l - 1$, $\w$ for $\li = \l - 1$, and $\o$ for $\li = \l$; the total primitive output size is $\y \sim \t\d\w + \o$. Finally, the forward pass costs $\fp \sim \t\d\f\w^2 + \o\w$ time and $\d\w + \f\w^2 + \o\w$ memory.
    
        As in \sref{sec:example_fcn} we plug the above into the cost of the \jc{} in \sref{sec:vanilla}, and obtain
        \begin{tcolorbox}
            CNN \jc{} costs 
            $\n^2\o^2\t\f\w^2 + \n^2\o^3\w + \n\o\t\d\f\w^2$ time; 
            $\n^2\o^2+\n\o\f\w^2+\n\t\d\w+\n\o^2\w+\t\f\w^2$ memory.
        \end{tcolorbox}
        Analogously, the cost of \ntvp{} from \sref{sec:implicit} becomes
        \begin{tcolorbox}
            CNN \ntvp{} cost 
            $\n^2\o\t\d\f\w^2 + \n^2\o^2\w$ time; 
            $\n^2\o^2+\n\o\f\w^2+\n\t\d\w+\n\o^2\w+\t\f\w^2$ memory.
        \end{tcolorbox}
        For \sd{} (\sref{sec:str_derivatives}), we remark that \textbf{MJJMP} and $\j$ costs for $\ti = \t$ are identical to FCN (\sref{sec:example_fcn}, \sref{sec:example_fcn_mjjmp}), hence $\n^2\o^3 + \n^2\w$ time and $\n^2\o^2 + \n\w$ memory. 
        
        For convolutional layers $\ti < \t$, we reference \sref{sec:example_cnn_mjjmp} and specifically \cref{tab:complexities_cnn}, to obtain $\j = \d\f\w$ for both memory and time, since the primitive Jacobian subarrays can be computed via $\f$ calls of \code{np.roll} on the $\d\times\w$ input $x$.  However, note that on actual hardware such operations are known to perform {\em worse} than highly optimized convolution-based solutions that scale as $\d\f^2\w$ (this would be performing $\f$ JVP calls of the convolutional primitive, each costing $\d\f\w$ time, or, equivalently, invoking \code{jax.lax.conv\_general\_dilated\_patches}). For this reason we currently use the convolutional approach in our implementation, but note that it would be trivial to extend the codebase to switch to the \code{np.roll}-based solution in the case of very large filters $\f > \o\w$. 
        
        Finally, from \cref{tab:complexities_cnn}, $\textbf{MJJMP} = \n^2\o^2 \min\left(\f\w^2, \d\w + \frac{\d\f\w^2}{\o}, \d\w + \frac{\d^2\w}{\o} + \frac{\d^2\f\w}{\o^2}\right)$ (we have ignored the $\n\o\d\f\w^2$ term in \textbf{Outside-in} contraction order since this is dominated by the cost  $\n\o\left[\fp\right]$ of computing primitive output cotangents). 

        Adding all the costs up, we obtain
        \begin{tcolorbox}
            CNN \sd{} cost \\
            $\n\o\t\d\f\w^2 + \n\o^2\w + \n^2\o^3 + \n^2\o^2\t \min\left(\f\w^2, \d\w + \frac{\d\f\w^2}{\o}, \d\w + \frac{\d^2\w}{\o} + \frac{\d^2\f\w}{\o^2}\right)$ time; \\
            $\n^2\o^2 + \n\o\d\w + \n\d\f\w + \n\t\d\w +\t\f\w^2 + \o\w^2$ memory.
        \end{tcolorbox}\vspace{-0.2cm}

    \subsection{Jacobian Rules for Structured Derivatives}\label{sec:jacobian_rules}
        
        Here we discuss computing primitive $\partial y / \partial \theta$ Jacobians as part of our \sd{} implementation in \sref{sec:implementation}. We provide 4 options to compute them through arguments \code{\_j\_rules} and \code{\_fwd}:
        \begin{enumerate}
            \item \textbf{Forward mode}, \code{\_fwd = True}, is equivalent to \code{jax.jacfwd}, forward mode Jacobian computation, performed by applying the JVP to $\p$ columns of the $I_\p$ identity matrix. Best for $\p < \y$.
            \item \textbf{Reverse mode}, \code{\_fwd = False}, is equivalent to \code{jax.jacrev}, reverse mode Jacobian computation, performed by applying the VJP to $\y$ columns of the $I_\y$ identity matrix. Best for $\p > \y$.
            \item \textbf{Automatic mode}, \code{\_fwd = None}, selects forward or reverse mode for each $y$ based on parameters and output shapes.
            \item \textbf{Rule mode}, \code{\_j\_rules = True}, queries a dictionary of Jacobian rules (similar to the dictionary of structure rules) with our custom implementations of primitive Jacobians, instead of computing them through VJPs or JVPs. The reason for introducing custom rules follows our discussion in \sref{sec:str_derivatives}: while JAX has computationally optimal VJP and JVP rules, the respective Jacobian computations are not guaranteed to be most efficient. In practice, we find our rules to be most often faster, however this effect is not perfectly consistent (can occasionally be slower) and often negligible,  requiring further investigation.
        \end{enumerate}
        The default setting is \code{\_j\_rules = True}, \code{\_fwd = None}, i.e. a custom Jacobian implementation is preferred, and, if absent, Jacobian is computed in forward or reverse mode based on parameters and output sizes. Note that in all settings, structure of $\partial y / \partial \theta$ is used to compute only the smallest Jacobian subarray necessary, and therefore most often inputs to VJP/JVP will be smaller identity matrices $I_{\p/\c}$ or $I_{\y/\c}$ respectively, and all methods will return a smaller Jacobian matrix of size $\left(\y/\c\right)\times\left(\p/\c\right)$. If for any reason (for example debugging) you want the whole $\partial y/\partial \theta$ Jacobians computed, you can set the \code{\_s\_rules=False}, i.e. disable structure rules.

    \subsection{Known Issues}\label{sec:known_issues}
        
        We will continue improving our function transformations in various ways after release, and welcome bug reports and feature requests. Below are the missing features / issues at the time of submission:
        \begin{enumerate}
            \item No support for complex differentiation.
            \item Not tested on functions with advanced JAX primitives like parallel collectives (\code{psum}, \code{pmean}, etc.), gradient checkpointing (\code{remat}), compiled loops (\code{scan}; Python loops are supported).
            \item Our current implementation of \ntvp~relies on \xla's common subexpression elimination (CSE) in order to reuse computation across different pairs of inputs $x_1$ and $x_2$, and, as shown in \cref{fig:fcn_flops} and \cref{fig:fcn_secs_other}, can have somewhat unpredictable wall-clock time performance and memory requirements. We believe this could correspond to CSE not always working perfectly, and are looking into a more explicitly efficient implementation.
        \end{enumerate}

    \subsection{Finite and Infinite Width NTK}\label{sec:finite_vs_inf}
        
        In this work we focus on the finite width NTK $\Theta^f_\theta\left(x_1, x_2\right)$, defined in \cref{eq:ntk_outer_product}, repeated below with a batch size $\n$:
        \begin{equation}\label{eq:ntk_outer_product_batched}
            \textbf{F-NTK (finite width):}\quad\underbrace{\Theta^f_\theta(x_1, x_2)}_{\n\o\times \n\o} \coloneqq \underbrace{\left[\partial f(\theta, x_1)\big/\partial \theta\right]}_{\n\o \times \p} \underbrace{\left[\partial f(\theta, x_2)\big/\partial \theta\right]^T}_{\p\times \n\o}.
        \end{equation}
        Another important object in deep learning is the \textit{infinite width} NTK $\Theta^f\left(x_1, x_2\right)$, introduced by \citet{Jacot2018ntk}:
        \begin{equation}\label{eq:inf_ntk_outer_product_batched}
            \textbf{I-NTK (infinite width):}\quad\underbrace{\Theta^f(x_1, x_2)}_{\n\o\times \n\o} \coloneqq \lim_{\w\to\infty}\mathbb{E}_{\theta\sim\mathcal{N}\left({\bf 0},I_{\p}\right)}\left[\underbrace{\Theta^f_\theta\left(x_1, x_2\right)}_{\n\o \times \n\o}\right].
        \end{equation}
        A natural question to ask is what are the similarities and differences of F- and I-NTK, when is one more applicable than the other, and what are their implementation and compute costs.
        
        \textbf{Applications.} At a high level, F-NTK describes the local/linearized behavior of the finite width NN $f\left(\theta, x\right)$ \citep{lee2019wide}. In contrast, I-NTK is an approximation that is exact only in the infinite width $\w$ limit, and only at initialization $\left(\theta\sim\mathcal{N}\left({\bf 0},I_{\p}\right)\right)$. As such, the resulting I-NTK has no notion of width $\w$, parameters $\theta$, and cannot be computed during draining, or in a transfer or meta-learning setting, where the parameters $\theta$ are updated. As a consequence, any application to finite width networks (\sref{sec:related}) is better served by the F-NTK, and often impossible with the I-NTK.
        
        In contrast, I-NTK describes the behavior of an infinite ensemble of infinitely wide NNs. In certain settings this can be desirable, such as when studying the inductive bias of certain NN architectures \citep{xiao2019disentangling} or uncertainty \citep{adlam2020exploring}, marginalizing away the dependence on specific parameters $\theta$. However, care should be taken when applying I-NTK findings to the finite width realm, since many works have demonstrated substantial finite width effects that cannot be captured by the I-NTK \citep{novak2018bayesian, arora2019fine, lee2019wide, yaida2019non, Hanin2020Finite, lee2020finite}.
        
        \textbf{Mathematical scope.} Another significant difference between F- and I-NTK is the scope of their definitions in \cref{eq:ntk_outer_product_batched} and \cref{eq:inf_ntk_outer_product_batched} and mathematical tractability. 
        
        The F-NTK is well-defined for any differentiable (w.r.t. $\theta$) function $f$, and so are our algorithms. In fact, our implementations support any Tangent Kernels (not necessarily ``Neural''), and are not specific to NNs at all.
        
        In contrast, the I-NTK requires the function $f$ to have the concept of width $\w$ (that can be meaningfully taken to infinity) to begin with, and further requires $f$ and $\theta$ to satisfy many conditions in order for the I-NTK to be well-defined \citep{yang2019scaling}. In order for I-NTK to be well-defined \textit{and computable in closed-form}, $f$ needs to be built out of a relatively small, hand-selected number of primitives that admit certain Gaussian integrals to have closed-form solutions. Examples of ubiquitous primitives that \textit{don't} allow a closed-form solution include attention with standard parameterization \citep{hron2020}; max-pooling; sigmoid, (log-)softmax, tanh, and many other nonlinearities; various kinds of normalization \citep{yang2018a}; non-trivial weight sharing \citep{yang2020tensor}; and many other settings. Going forward, it is unlikely that the I-NTK will scale to the enormous variety of architectures introduced by the research community each year.
    
        \textbf{Implementation tractability.} Above we have demonstrated that the I-NTK is defined for a very small subset of functions admitting the F-NTK. A closed-form solution exists for an even smaller subset. However, even when the I-NTK admits a closed-form solution, it is important to consider the complexity of implementing it.
        
        Our implementation for computing the F-NTK is applicable to any differentiable function $f$, and requires no extra effort when switching to a different function $g$. It is similar to JAX's highly-generic function transformations such as \code{jax.vmap}.
        
        In contrast, there is no known way to compute the I-NTK for an arbitrary function $f$, even if the I-NTK exists in closed form. The best existing solution to date is provided by \citet{neuraltangents2020}, which allows to \textit{construct} $f$ out of the limited set of building blocks provided by the authors. However, one cannot compute the I-NTK for a function implemented in a different library such as Flax \citep{flax2020github}, or Haiku \citep{haiku2020github}, or bare-bone JAX. One would have to re-implement it using the primitives provided by \citet{neuraltangents2020}. Further, for a generic architecture, the primitive set is unlikely to be sufficient, and the function will need to be adapted to admit a closed-form I-NTK.
    
        \textbf{Computational tractability.} F-NTK and I-NTK have different time and memory complexities, and a fully general comparison is an interesting direction for future work. Here we provide discussion for deep FCNs and CNNs.
        
        Networks having a fully-connected top ($\t$) readout layer have a constant-block diagonal I-NTK, hence its cost \textit{does not} scale with $\o$. The cost of computing the I-NTK for a deep FCN scales as $\n^2\t$ for time and $\n^2$ for memory. A deep CNN without pooling costs $\n^2\d\t$ time and $\n^2\d$ memory (where $\d$ is the total  number of pixels in a single input/activation; $\d = 1$ for FCNs). Finally, a deep CNN with pooling, or any other generic architecture that leverages the spatial structure of inputs/activations, costs $\n^2\d^2\t$ time and $\n^2\d^2$ memory. This applies to all models in \cref{fig:resnet_secs} and \cref{fig:imagenet_other_secs}, Graph Neural Networks \citep{du2019graph}, and the vast majority of other architectures used in practice.
        
        The quadratic scaling of the I-NTK cost with $\d$ is especially burdensome, since, for example, for ImageNet $\d^2 = 224^4 = 2,517,630,976$. As a result, it would be impossible to evaluate the I-NTK on even a single ($\n = 1$) pair of inputs with a V100 GPU for any model for which we've successfully evaluated the F-NTK in \cref{fig:resnet_secs} and \cref{fig:imagenet_other_secs}.
        
        The F-NTK time and memory only scale linearly with $\d$ (\cref{tab:intro_summary_1x1}). However, the F-NTK cost scales with other parameters such as width $\w$ or number of outputs $\o$, and in general the relative F- and I-NTK performance will depend on these parameters. As a rough point of comparison, we consider the cost of evaluating the I-NTK of a 20-layer binary classification ReLU CNN with pooling on a V100 GPU used by \citet{arora2019fine} against the respective F-NTK with $\w = 128$ also used by \citet[Section B]{arora2019fine}. \citet{arora2019fine} and \citet{neuraltangents2020} report from $0.002$ to $0.003$ seconds per I-NTK entry on a pair of CIFAR-10 inputs. Using \sd, we can compute the respective F-NTK entry on same hardware in at most $0.000014$ seconds, i.e. at least 100 times faster than the I-NTK. In $0.002$ -- $0.003$ seconds per NTK entry, we can compute the F-NTK on a pair of \textit{ImageNet} inputs (about 50x larger than CIFAR-10) for a \textit{200-layer ResNet} (about 10x deeper than the model above) in \cref{fig:resnet_secs} (top left).
        
        Finally, we remark that efficient NTK-vector products without instantiating the entire $\n\o\times\n\o$ NTK are only possible using the F-NTK (\sref{sec:ntk_vp_wo_ntk}).

    \subsection{Relationship Between the NTK and the Hessian}\label{sec:ntk_hessian}
        Here we briefly touch on the difference between the NTK 
        \begin{align}
            \textbf{NTK:}\quad\,\,\underbrace{\Theta^f_\theta(x_1, x_2)}_{\n\o\times \n\o} &\coloneqq \underbrace{\left[\partial f(\theta, x_1)\big/\partial \theta\right]}_{\n\o \times \p} \underbrace{\left[\partial f(\theta, x_2)\big/\partial \theta\right]^T}_{\p\times \n\o},\\
            \intertext{and the Hessian:}
            \textbf{Hessian:}\quad\quad\underbrace{{\bf H}_\theta (x)}_{\p\times \p} &\coloneqq \frac{\partial^2 \mathcal{L}\left(f\left(\theta, x\right)\right)}{\partial \theta^2},
        \end{align}
        defined for some differentiable loss function on the output space $\mathcal{L}\colon\mathbb{R}^{\n\o}\to\mathbb{R}$. 
        
        Both matrices characterize localized training dynamics of a NN, and the NTK can be used as a more tractable quantity in cases where the Hessian is infeasible to instantiate (for example, $\p$ amounts to tens of millions in models considered in \cref{fig:resnet_secs} and \cref{fig:imagenet_other_secs}).
        
        The connection between the NTK and the Hessian can be established for when $\mathcal{L}$ is the squared error (SE), i.e. $\mathcal{L}(y) = \left\|y - \mathcal{Y}\right\|_2^2 / 2$, where $\mathcal{Y}\in\mathbb{R}^{\n\o}$ are the training targets. In this case, as presented in \citep[Section 2]{pmlr-v70-pennington17a} and \citep[Equation 13; page 21]{grosse2021neural}:
        \begin{equation}
            \underbrace{{\bf H}_\theta (x)}_{\p\times \p} = \underbrace{\left[\frac{\partial f(\theta, x)}{\partial \theta}^T \frac{\partial f(\theta, x)}{\partial \theta}\right]\vphantom{\vspace{1cm}}}_{\coloneqq{\bf H}^0_\theta (x)} + \underbrace{\sum_{\ni,\oi = 1}^{\n,\o}\left(f\left(\theta, x\right) - \mathcal{Y}\right)^{\ni, \oi}\frac{\partial^2 f\left(\theta, x\right)^{\ni,\oi}}{\partial^2\theta}}_{\coloneqq{\bf H}^1_\theta (x)},
        \end{equation}
        where we have decomposed the Hessian ${\bf H}_\theta (x)$ into two summands ${\bf H}^0_\theta (x)$ and ${\bf H}^1_\theta (x)$ following the notation of \citet{pmlr-v70-pennington17a}.
        
        Notice that if $f\left(\theta, x\right) = \mathcal{Y}$, i.e. the SE loss is $0$, ${\bf H}^1_\theta (x) = 0$, yielding
        
        \begin{equation}
            \underbrace{{\bf H}_\theta (x)}_{\p\times \p} = {\bf H}^0_\theta (x) = \underbrace{\frac{\partial f(\theta, x)}{\partial \theta}^T}_{\p\times\n\o} \underbrace{\frac{\partial f(\theta, x)}{\partial \theta}}_{\n\o\times\p},\quad\quad\quad\quad
            \underbrace{\Theta^f_\theta(x, x)}_{\n\o\times \n\o} = \underbrace{\frac{\partial f(\theta, x)}{\partial \theta}}_{\n\o \times \p} \underbrace{\frac{\partial f(\theta, x_2)}{\partial \theta}^T}_{\p\times\n\o},
        \end{equation}
        and, as a consequence, the Hessian and the NTK have the same eigenvalues (see also \citet[Page 21]{grosse2021neural}) in this particular case. Moreover, the Hessian (and Hessian-vector products) can be computed very similarly to \ntvp, by switching the order of VJP and JVP operations in \cref{eq:ntk_vp_main}.
        
        However, except for zero SE loss case above, the NTK and the Hessian have different spectra, and their computations share less similarity. Precisely, Hessian-vector products (and consequently the Hessian) are computed in JAX through a composition of JVPs and VJPs similar to \ntvp:
        \begin{align}\label{eq:hvp}
            {\bf H}_\theta (x)v &= \left[\frac{\partial^2 \mathcal{L}\left(f\left(\theta, x\right)\right)}{\partial \theta^2}\right]v = \frac{\partial}{\partial \theta}\left[\frac{\partial \mathcal{L}\left(f\left(\theta, x\right)\right)}{\partial \theta}\right]v = \frac{\partial \left[\textrm{VJP}_{\left(\theta, x\right)}^{\mathcal{L}\circ f}\left(1\right)\right]}{\partial \theta} v = \textrm{JVP}^{\left[\textrm{VJP}^{\mathcal{L}\circ f}_{\left(\cdot, \cdot\right)}\left(1\right)\right]}_{\left(\theta, x\right)}\left(v\right).
        \end{align}
        Although \cref{eq:hvp} is similar to \cref{eq:ntk_vp_main} in that both are compositions of JVPs and VJPs, in \cref{eq:ntk_vp_main} the result of a VJP is the input tangent to the JVP of $f$, while in \cref{eq:hvp} it is the function to be differentiated by the JVP (instead of $f$).

    \subsection{Relationship Between \texorpdfstring{\sd}{Structured Derivatives} and K-FAC}\label{sec:additional_derivation}
        
        Similarly to K-FAC \citep{martens2015optimizing}, in the simple example of \sref{sec:str_derivatives}, we leverage the structure in the matrix-vector product derivative w.r.t. the matrix, and we use the mixed-product property, i.e. $\left(A\otimes B\right)\left(C\otimes D\right) = \left(A C\right)\otimes\left(B D\right)$. However, in the general case this is not enough, and \sd~rely on three components: (1) a direct sum linear algebra \cref{eq:direct_sum}, (2) symbolic simplification of expressions with an identity matrix (\sref{sec:constant_block_diagonal}), and (3) optimal order of contractions in \cref{eq:expanded_ntk_main} (e.g. ``Inside-out'' (\cref{tab:complexities_fcn}), which is not possible to achieve with standard AD tools). All three components are necessary to achieve our asymptotic complexities, and cannot be achieved by leveraging the mixed-product property alone.

    \subsection{Applications with a Limited Compute Budget}\label{sec:ntk_vp_wo_ntk}
        While our methods allow to dramatically speed-up the computation of the NTK, all of them still scale as $\n^2\o^2$ for both time and memory, which can be intractable for large datasets and/or large outputs.
        
        Here we present several settings in which our proposed methods still provide substantial time and memory savings, even when instantiating the entire $\n\o\times\n\o$ NTK is not feasible or not necessary.
        
        \begin{itemize}
            \item \textbf{NTK-vector products}. In many applications one only requires computing the NTK-vector product linear map
            \begin{equation}\label{eq:ntvp_supp}
                \Theta_\theta^f\colon v \in \mathbb{R}^{\n\o}\mapsto \Theta_{\theta}^f v \in \mathbb{\n\o},
            \end{equation}
            without computing the entire NTK matrix $\Theta_\theta$. A common setting is using the power iteration method \citep{powermethod} to compute the NTK condition number and hence trainability of the respective NN \citep{lee2019wide, chen2020tenas, chen2021vision}. Another setting is using conjugate gradients to compute $\Theta_{\theta}^{-1}\mathcal{Y}$ when doing kernel ridge regression with the NTK \citep{Jacot2018ntk, lee2019wide, zhou2021metalearning}.
            
            \cref{eq:ntvp_supp} is the same map as the one we considered in \sref{sec:implicit}, and naturally, \ntvp~can provide a substantial speed-up over \jc~in this setting. Precisely, a straightforward application of \jc~yields
            \begin{equation}
                \underbrace{\Theta^f_{\theta}v}_{\n\o\times 1} 
                =\underbrace{\frac{\partial f\left(\theta, x_1\right)}{\partial\theta}}_{\n\o\times \p}\quad\underbrace{\frac{\partial f\left(\theta, x_2\right)}{\partial\theta}^T}_{\p \times \n\o}\quad \underbrace{v}_{\n\o\times 1}.
            \end{equation}
            Combined with the cost of computing the weight space cotangents $\partial f/\partial \theta$, such evaluation costs $\n\o\left[\fp\right]$ time, i.e. the cost of instantiating the entire \jac. Alternatively, one could store the entire Jacobians of sizes $\n\o\p$ in memory, and compute a single NTK-vector product in $\n\o\p$ time.
            
            In contrast, \ntvp~allow to compute an NTK-vector product at a cost asymptotically equivalent to a single VJP call (\sref{sec:implicit}), i.e. $\n\left[\fp\right]$, $\o$ times faster than \jc~without caching. With caching, fastest method will vary based on the cost of $\left[\fp\right]$ relative to $\o\p$, as discussed in \sref{sec:implicit}, but \ntvp~will remain substantially more memory-efficient due to not caching the entire $\n\o\p$ Jacobians.
            
            \item \textbf{Batching.} In many applications it suffices to compute the NTK over small batches of the data. For example \citet{dauphin2019metainit, chen2020tenas, chen2021vision} estimate the conditioning by computing an approximation to the NTK on $\n$ equal to 128, 32, and 48 examples respectively. Similarly, \citet{zhou2021metalearning} use a small batch size of $\n = 25$ to meta-learn the network parameters by replacing the inner SGD training loop with NTK regression.
            
            \item \textbf{Pseudo-NTK.} Many applications (\sref{sec:related}) compute a pseudo-NTK of size $\n\times\n$, which is commonly equal to one of its $\o$ diagonal blocks, or to the mean of all $\o$ blocks. The reason for considering such approximation is that in the infinite width limit, off-diagonal entries often converge to zero, and for wide-enough networks this approximation can be justified. Compute-wise, these approximations are equivalent to having $\o = 1$. While an important contribution of our work is to enable computing the full $\n\o\times \n\o$ NTK quickly, if necessary, \sd~can be combined with the $\o = 1$ approximations, and still provide an asymptotic speed-up and memory savings relative to prior works.
            
        \end{itemize}

    \subsection{Leveraging JAX Design for Efficient NTK Computation}\label{sec:jax}
        
        At the time of writing, Tensorflow \citep[TF]{abadi2016tensorflow} and PyTorch \citep{pytorch} are more widely used than JAX \citep{jax2018github}. However, certain JAX features and design choices made it much more suitable, if not indispensable, for our project:
        \begin{enumerate}
            \item \sd~require manual implementation of structure rules for different  primitives in the computational graph of a function $f(\theta, x)$. JAX has a small primitive set of about 136 primitives, while PyTorch and Tensorflow have more than 400 \citep[Section 2]{frostig2021decomposing}. Further, by leveraging \code{jax.linearize}, we reduce our task to implementing structure rules for only \textit{linear} primitives, of which JAX has only 56.\footnote{See \sref{sec:str_derivatives}, as well as \citep[Section 1]{frostig2021decomposing} for how JAX uses the same insight to not implement all 136 VJP rules, but only implement 56 transpose rules for reverse mode AD.} To our knowledge neither PyTorch nor Tensorflow have an equivalent transformation, which makes JAX a natural choice due to the very concise set of primitives that we need to handle (\cref{tab:list_of_primitives}).
            
            \item \ntvp~critically rely on forward mode AD (JVP), and \sd~also use it (albeit it’s not crucial; see \sref{sec:jacobian_rules}). At the time of writing, PyTorch \href{https://pytorch.org/docs/stable/generated/torch.autograd.functional.jvp.html#torch.autograd.functional.jvp}{does not implement an efficient forward mode AD}.
            
            \item \sd~rely crucially on the ability to traverse the computation graph to rewrite contractions using our substitution rules. JAX provides a highly-convenient graph representation in the form of a  \href{https://jax.readthedocs.io/en/latest/jaxpr.html}{Jaxpr}, as well as \href{https://jax.readthedocs.io/en/latest/notebooks/Writing_custom_interpreters_in_Jax.html}{tooling and documentation} for writing custom Jaxpr interpreters.
            
            \item All implementations (even \jc) rely heavily on \code{jax.vmap} (and in many cases, it is indispensable). While PyTorch has \href{https://github.com/pytorch/pytorch/issues/42368}{released} a prototype of \code{vmap} in May 2021, it was not available when we started this project.
        \end{enumerate}

        \subsubsection{Interfacing with Tensorflow}
            Since JAX and TF leverage the same underlying compiler \href{https://www.tensorflow.org/xla}{XLA}, we are able to construct a seamless TF $\to$ JAX $\to$ TF pipeline using \href{https://github.com/google/jax/tree/main/jax/experimental/jax2tf}{Jax2TF} and \href{https://github.com/google/jax/tree/main/jax/experimental/jax2tf}{TF2Jax} \citep{deepmind2020jax}. We will provide a prototype implementation of this pipeline in the \code{experimental} folder.

        \subsubsection{Interfacing with PyTorch}
            Recently introduced Functorch \citep{functorch2021} enables implementing \jc{} and \ntvp{}, and our code has been ported to PyTorch in the \href{https://pytorch.org/functorch/stable/notebooks/neural_tangent_kernels.html}{``Neural Tangent Kernels'' tutorial}. However, due to JAX features (1) and (3) from \sref{sec:jax} implementing \sd{} in PyTorch remains challenging. If necessary, a pipeline PyTorch $\to$ TF $\to$ JAX $\to$ PyTorch using \href{https://onnx.ai/}{ONNX} \citep{bai2019} and \href{https://github.com/dmlc/dlpack}{DLPack} can be constructed. We will include an example implementation in the \code{experimental} folder.

    \subsection{Experimental Details}\label{sec:experimental_definition}

        All experiments were performed in JAX \citep{jax2018github} using 32-bit precision.
        
        Throughout this work we assume the cost of multiplying two matrices of shapes $(M, K)$ and $(K, P)$ to be $M K P$. While there are \href{https://en.wikipedia.org/wiki/Matrix_multiplication_algorithm}{faster algorithms} for very large matrices, the \xla~compiler (used by JAX, among other libraries) does not implement them, so our assumption is accurate in practice.
        
        \textbf{Hardware.} \textbf{CPU} experiments were run on Dual 28-core Intel Skylake CPUs with at least 240 GiB of RAM. \textbf{NVIDIA V100} and \textbf{NVIDIA P100} used a respective GPU with 16 GiB GPU RAM. \textbf{TPUv3} and \textbf{TPUv4} have 8 and 32 GiB of RAM respectively, and use the default 16/32-bit mixed precision.
        
        \cref{fig:fcn_flops} and \cref{fig:fcn_secs_other}: a 10-layer, ReLU FCN was constructed with the Neural Tangents \citep{neuraltangents2020} \code{nt.stax} API. Defeault settings (weight variance $1$, no bias) were used. Individual inputs $x$ had size 3. \jc~was evaluated using \code{nt.empirical\_ntk\_fn} with \code{trace\_axes=(), diagonal\_axes=(), vmap\_axes=0}. \jac{} was evaluated using \code{jax.jacobian} with a \code{vmap} over inputs $x$. For time measurements, all functions were \code{jax.jit}ted, and timing was measured as the average of 100 random samples (compilation time was not included). For FLOPs, the function was not JITted, and FLOPs were measured on CPU using the \code{utils.get\_flops} function that is released together with our code.\footnote{The \xla~team has let us know that if JITted, the FLOPs are currently correctly computed only on TPU, but are incorrect on other platforms. Therefore we compute FLOPs of non-JITted functions.}
        
        \cref{fig:resnet_secs} and \cref{fig:imagenet_other_secs}: for ResNets, implementations from Flax \citep{flax2020github} were used, specifically \code{flax.examples.imagenet.models}. For WideResNets, the \href{flax.examples.imagenet}{code sample} from \citet{neuraltangents2020} was used.\footnote{We replaced \code{stax.AvgPool((8, 8)), stax.Flatten()} with \code{stax.GlobalAvgPool()}.} For all other models, we used implementations from \url{https://github.com/google-research/vision_transformer}. Inputs were random arrays of shapes $224 \times 224 \times 3$. All models were JITted. All reported values are averages over 10 random samples. For each setting, we ran a grid search over the batch size $\n$ in $\left\{2^k\right\}_{k=0}^{9}$, and reported the best time divided by $\n^2$, i.e. best possible throughput in each setting.
        
        Title page ribbon is adapted from \citet{vecteeze}.

\end{document}